\documentclass[10pt,twocolumn,letterpaper]{article}

\usepackage[pagenumbers]{wacv} 

\usepackage{graphicx}
\usepackage{amsmath}
\usepackage{amssymb}
\usepackage{appendix}
\usepackage{booktabs}
\usepackage{multirow} 
\usepackage{pgfplots}
\pgfplotsset{compat=1.18}
\usepgfplotslibrary{groupplots}
\usepackage[none]{hyphenat}
\usepackage{tikz}
\usepackage{pgfplots}
\pgfplotsset{compat=1.18} 
\usepackage{pifont}
\usepackage{dsfont}
\usepackage{algorithm}
\usepackage{algpseudocode}

\usepackage[pagebackref,breaklinks,colorlinks]{hyperref}

\usepackage[capitalize]{cleveref}
\crefname{section}{Sec.}{Secs.}
\Crefname{section}{Section}{Sections}
\Crefname{table}{Table}{Tables}
\crefname{table}{Tab.}{Tabs.}

\def\wacvPaperID{207} 
\def\confName{WACV}
\def\confYear{2025}

\begin{document}

\title{Uni-SLAM: Uncertainty-Aware Neural Implicit SLAM for Real-Time Dense Indoor Scene Reconstruction}

\author{Shaoxiang Wang, Yaxu Xie, Chun-Peng Chang, Christen Millerdurai, Alain Pagani, Didier Stricker\\
German Research Center for Artificial Intelligence \\
{\tt\small firstname.lastname@dfki.de}
}

\maketitle

\begin{abstract}
   Neural implicit fields have recently emerged as a powerful representation method for multi-view surface reconstruction due to their simplicity and state-of-the-art performance. However, reconstructing thin structures of indoor scenes while ensuring real-time performance remains a challenge for dense visual SLAM systems. Previous methods do not consider varying quality of input RGB-D data and employ fixed-frequency mapping process to reconstruct the scene, which could result in the loss of valuable information in some frames.

    In this paper, we propose Uni-SLAM, a decoupled 3D spatial representation based on hash grids for indoor reconstruction. We introduce a novel defined predictive uncertainty to reweight the loss function, along with strategic local-to-global bundle adjustment. Experiments on synthetic and real-world datasets demonstrate that our system achieves state-of-the-art tracking and mapping accuracy while maintaining real-time performance. It significantly improves over current methods with a 25\% reduction in depth L1 error and a 66.86\% completion rate within 1 cm on the Replica dataset, reflecting a more accurate reconstruction of thin structures. Project page: \href{https://shaoxiang777.github.io/project/uni-slam/}{https://shaoxiang777.github.io/project/uni-slam/}

\end{abstract}

\section{Introduction}
\label{sec:intro}

\begin{figure}[t]
\captionsetup{skip=3pt}
    \centering
    \subcaptionbox{ \label{fig:woUnc}}[0.45\linewidth]{
        \includegraphics[width=\linewidth]{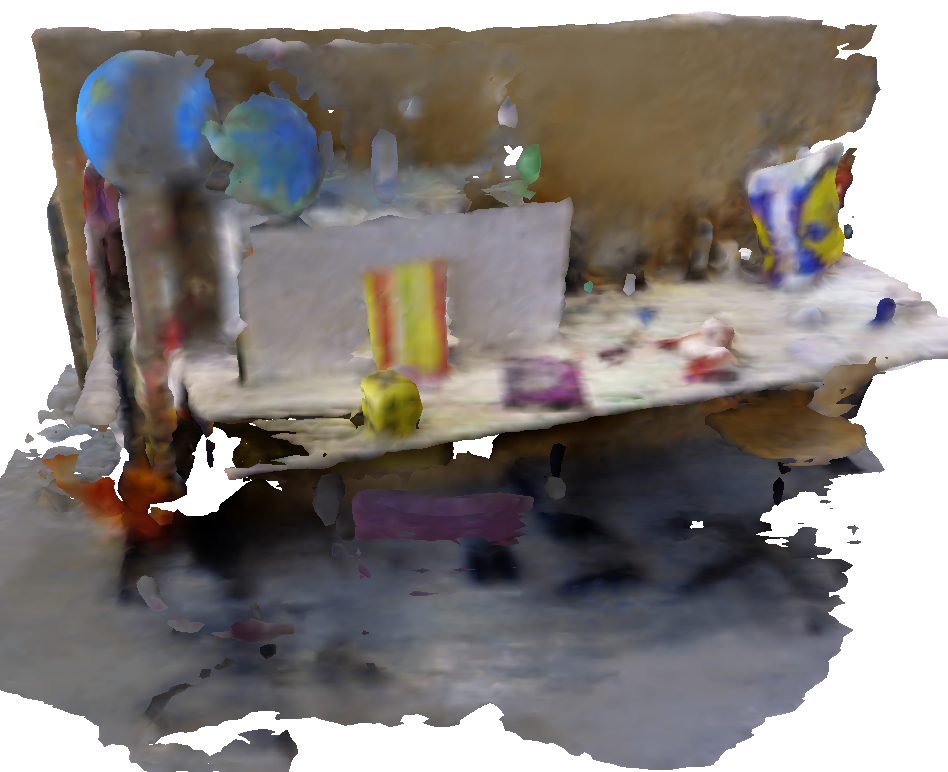}
    }
    \hfill
    \subcaptionbox{ \label{fig:withUnc}}[0.45\linewidth]{
        \includegraphics[width=\linewidth]{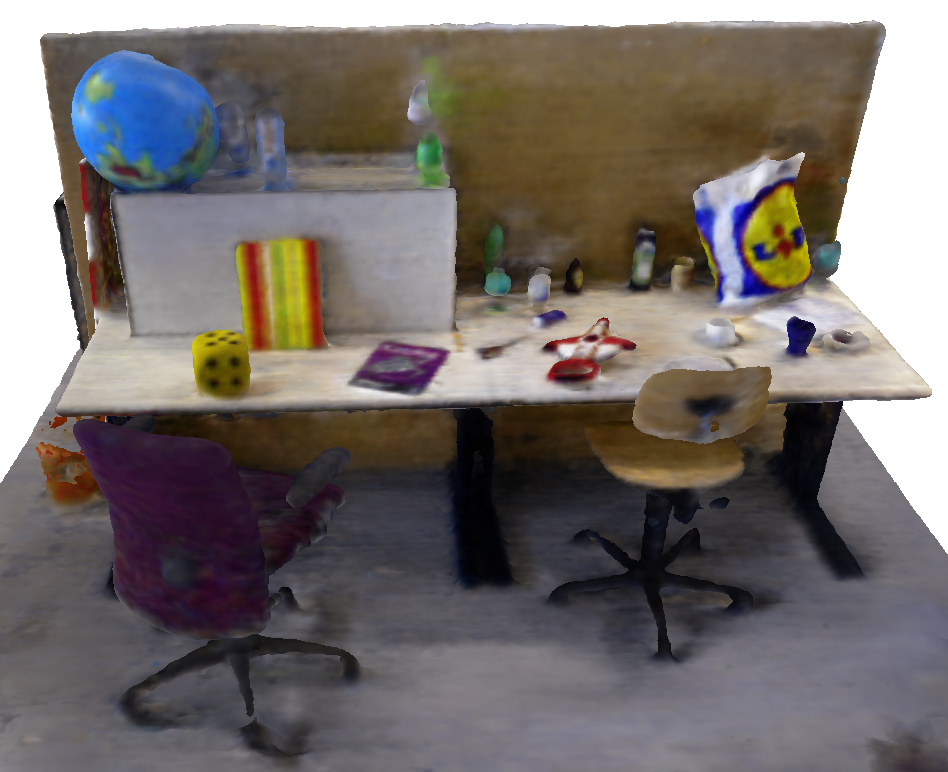}
    }
    \caption{ 
    The reconstructed 3D mesh on the TUM RGB-D dataset \cite{tumrgbd}, generated using our proposed method without uncertainty-guided reweighting and strategy, is illustrated in \cref{fig:woUnc}. Conversely, \cref{fig:withUnc} demonstrates the 3D mesh produced by our method after the incorporation of the uncertainty-aware strategy.}
    \label{fig:comparisonUnc}
\vspace{-5mm}
\end{figure}

Dense visual Simultaneous Localization and Mapping (SLAM) aims at reconstructing a dense 3D map of an unknown environment while simultaneously estimating the accurate camera pose. Traditional SLAM algorithms \cite{engel2014lsd, orbslam, dtam, svo} focus on localization accuracy for real-time large-scale applications, whereas Neural Radiance Fields (NeRFs) \cite{mildenhall2021nerf} significantly enhance dense 3D reconstruction and novel view synthesis, spurring the development of NeRF-based dense visual SLAM techniques.

As pioneering efforts, iMAP \cite{imap} and Nice-SLAM \cite{niceslam} utilize neural representations for both tracking and mapping, but slow convergence limits their low-latency mapping capabilities. SDF-based methods \cite{coslam,birn-slam, eslam, plgslam} offer faster convergence and higher rendering accuracy. But they treat all data even with varying quality equally, alternating tracking and mapping at a constant frequency (every $n$ frames). However, in dense NeRF-SLAM, the quality of RGB-D input data varies throughout the sequence (such as invalid depth), significantly impacting both camera pose estimation and scene reconstruction. Furthermore, constant mapping, this simple approach may lead to missing potentially effective information in frames where no mapping process occurs.  Therefore, treating all data uniformly in dense NeRF-SLAM systems is suboptimal, leading to overconfidence in poor-quality data and inefficient use of valuable information. 

Our dense SLAM method, Uni-SLAM, tackles these challenges by: 1) Differentiating data quality through pixel-level uncertainty analysis and loss reweighting to identify outliers; 2) Using image-level uncertainty to guide local-to-global bundle adjustment for comprehensive reconstruction; and 3) Employing decoupled hash grids to separately represent geometry and appearance, enabling real-time capture of high-frequency details in indoor scenes.

\noindent \textbf{Contributions} of our method are summarized as follows:
\begin{itemize}
\item 

We introduce a novel form of uncertainty, termed \textit{predictive uncertainty}, which enables pixel-level loss reweighting without the need for additional training. By leveraging this uncertainty, our method dynamically identifies and prioritizes valuable regions in the input data, enhancing the performance of mapping and tracking processes. This approach proves particularly effective when dealing with varying levels of input data quality, ensuring more robust and accurate outcomes.

\item 
Image-level uncertainty dynamically activates mapping with strategic local-to-global bundle adjustment, preserving valuable image information and enhancing global stability while capturing local color and geometry.

\item 
We propose an efficient scene representation using hash grids to decouple the scene's geometry and appearance. This approach enhances spatial representation of high-frequency signals while maintaining real-time performance. Our method achieves state-of-the-art results on the Replica \cite{replica}, ScanNet \cite{scannet}, and TUM RGB-D \cite{tumrgbd} datasets.

\end{itemize}

\section{Related Work}
The proposed method encompasses SLAM, implicit spatial representation and uncertainty modeling. Therefore, we focus the discussion of related work on these specific methods to better highlight our contributions.

\noindent\textbf{Dense Visual SLAM.}  Early dense visual SLAM approaches, like PTAM \cite{klein2007parallel} and DTAM \cite{dtam}, used feature-based methods, separating tracking and mapping tasks for efficiency. ORB-SLAM \cite{orbslam} further refined this with a feature-based approach for camera trajectory and 3D map construction. DROID-SLAM \cite{droidslam} introduced optical flow for precise real-time visual odometry and dense mapping. Learning-based methods \cite{yokozuka2019vitamin, saputra2020deeptio, li2020deepslam} improved feature extraction and robustness. Recent works \cite{mao2023ngel, nerfslam, goslam, orbeez, vmap} combine ORB-SLAM for robust tracking with NeRF-based mapping. Others \cite{imap, niceslam, coslam, pointslam, loopyslam, nicerslam, birn-slam, mipsFusion} integrate tracking and mapping in an interactive process. This paper explores uncertainty's impact in joint optimization scenarios.

\noindent\textbf{Scene Representation.}  Most common scene representation for dense mapping are grid-based (including voxel grids \cite{kinectfusion, 1996volumetric, neuralrecon}, octrees \cite{voxfusion,takikawa2021neural}, voxel hashing\cite{niessner2013real, instantngp}), surfel clouds\cite{pointnerf, elasticfusion, cho2021sp} and multi layer perceptron (MLP)-based \cite{neuralrgbd, nerfslam, yariv2021volume}. Grid-based methods offer the advantages of easy neighborhood finding and fast tri-linear interpolation. However, they require manual grid resolution specification and waste memory in empty regions \cite{niceslam,voxfusion,nicerslam}. Point-based methods avoid pre-specified resolutions but have complex neighborhood searches and low convergence speeds, which hinder real-time reconstruction. Additionally, they cannot fill in empty holes or make reasonable guesses for unscanned areas \cite{pointslam, pointnerf,gaussianslam1,gaussianslam2, gaussianslam3}. MLP-based methods suffer from slow convergence and catastrophic forgetting in large scenes \cite{imap,mipsFusion}, as updating all weights during optimization can cause forgetting issues.

\begin{figure*}[tb]
  \centering
  \includegraphics[height=6.5cm]{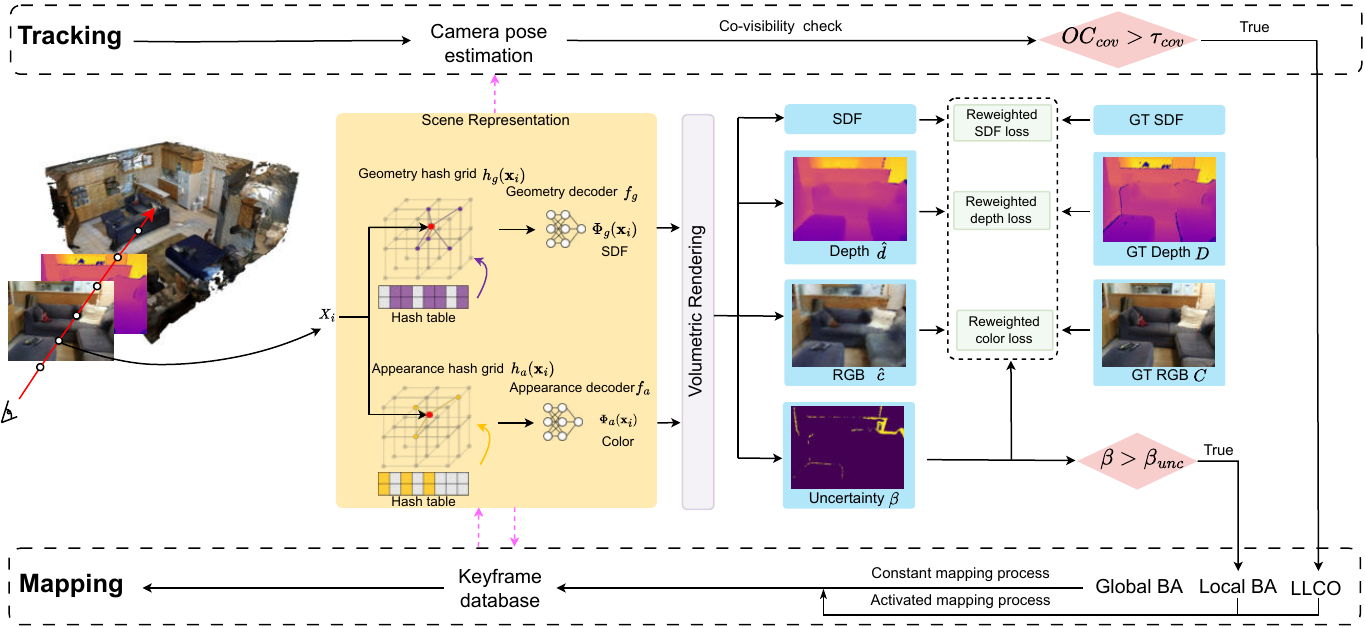}
  \caption{\textbf{Uni-SLAM Architecture Overview.} Our framework consists of two threads, tracking and mapping. While tracking is performed every frame for RGB-D stream, besides constant mapping is performed every $n$ frame constantly with global BA, activated additional mapping process is executed to capture local scene information based on uncertainty and co-visibility check with local BA and local loop closure optimization (LLCO). Our proposed pixel-level uncertainty method adaptively filters outlier pixels and reweights the loss function, enabling more precise localization during tracking and the reconstruction of color and geometric information in mapping.
 }
  \label{fig:activeSLAMPipeline}
\vspace{-5mm}
\end{figure*}

\noindent\textbf{Uncertainty Modeling in Scene Reconstruction.} The computer vision community has increasingly recognized the importance of uncertainty estimation across fields such as next-best-view (NBV) selection \cite{uncertaintyrobtotic3d, activenerf, sunderhauf2023density}, segmentation \cite{maag2024uncertainty, holder2021efficient, kahl2024values}, depth estimation \cite{hornauer2022gradient, hornauer2023out}, and SLAM \cite{uncleslam, carrillo2012comparison, chen2023pixel}. Uncertainty assessment enhances model interpretability and reduces critical errors. Kendall et al. \cite{uncertainties2017} identify two types of uncertainty in Bayesian deep learning: \emph{aleatoric} (due to inherent data ambiguity) and \emph{epistemic} (arising from limited data) \cite{ardeshir2022uncertainty, wursthorn2022comparison, hullermeier2021aleatoric}.

In NeRF-based novel view synthesis with \textit{known camera pose}, integrating uncertainty has led to improvements in handling blur, dynamic objects, and confidence visualization \cite{goli2024bayes, xue2024neural, shen2022conditional, martin2021nerf, shen2021stochastic}. However, its application in dense NeRF-SLAM with \textit{unknown camera pose} remains underexplored. Sandström \etal \cite{uncleslam} introduce a SLAM system that estimates aleatoric depth uncertainty, while Rosinol \etal \cite{rosinol2023probabilistic} propose fast uncertainty propagation for cleaner 3D meshes. To our knowledge, we are the first to use novel-defined predictive uncertainty, caused by limited unobserved data, to reweight dense implicit SLAM and guide local-to-global BA.

\section{Method}
Our overall pipeline is illustrated in \cref{fig:activeSLAMPipeline}. The input consists of a sequence of RGB-D images and known camera intrinsic parameters. Through a decoupled scene representation, we estimate the camera pose, the implicit truncated signed distance function (TSDF), depth, color and uncertainty. In \cref{sec:neuaralSceneRepresentation}, our efficient independent scene representation using two hash grids is described. In \cref{sect:uncertaintyModel}, we present our novel uncertainty model and explain how it reweights the loss function in \cref{sect:lossFunction}. Finally, \cref{sect:dynamicBA} presents the uncertainty-guided strategic BA and keyframe selection.

\subsection{Neural Scene Representation}
\label{sec:neuaralSceneRepresentation}
All existing implicit NeRF-based SLAM systems exhibit various issues
in scene representation: \emph {\textbf{a)} MLP-based\cite{imap} forgetting problem and insufficient spatial representation capability when using tri-planes\cite{chan2022efficient, eslam}. \textbf{b)} Coupled geometry and appearance information\cite{coslam,niceslam, plgslam} increases training difficulty, resulting in poor reconstruction quality. \textbf{c)} Coarse-to-fine dense grids\cite{niceslam} rely on heuristic resolution selection and require longer training times and high memory usage, failing to meet real-time requirements.} In our method, the hypothesis is that geometry and color information should not be sampled at the same frequency. To verify this, we opt for a decoupled representation, using multiresolution hash grids\cite{instantngp} model for each of them. We show in our experiments that this decoupled hash grid representation favors speed, hole-filling ability, and low memory footprint while not sacrificing accuracy. The raw SDF $\Phi_g(\mathbf{x}_i)$ and the raw color $\Phi_a(\mathbf{x}_i)$ are decoded via tiny MLPs geometry decoder $f_g$ and appearance decoder $f_a$:
\begin{equation}
\Phi_g(\mathbf{x}_i) = f_g\left(h_g(\mathbf{x}_i)\right) \quad \text{and} \quad \Phi_a(\mathbf{x}_i) = f_a\left(h_a(\mathbf{x}_i)\right)
\end{equation}
where $h_g(\mathbf{x}_i)$ and $h_a(\mathbf{x}_i)$ represent multiresolution geometry hash grids and appearance hash grids respectively in \cref{fig:activeSLAMPipeline}. We set the multiresolution level to $L = 16$, and only visualize one resolution level hash grid here for clarity. The decoupled representation effectively reduces the network's confusion when faced with appearance and geometry information of varying complexity. For more implementation details of hash grid, we refer readers to the supplementary Sec. A.1, B.1 and \cite{instantngp}.

\noindent\textbf{Depth and Color Volume Rendering.}  We follow \cite{niceslam} to render depth and color via integration along the sampling rays as $ \hat{\boldsymbol{c}}=\sum_{i=1}^N w_i \boldsymbol{\phi}_{\boldsymbol{a}}\left(\mathbf{x}_i\right) \quad $ and $\quad \hat{\boldsymbol{d}}=\sum_{i=1}^N w_i d_i $, where $d_i$ represents the distance from camera center to the current sample point $\mathbf{x}_i$ along this ray. $\mathbf{x}_i$ is sampled and guided by depth image as \cite{coslam}.  $w_i$ is the weight of the current sampling point, which can be converted from the density $ \sigma(\mathbf{x}_i) $ as 
\vspace{-2mm}
\begin{equation}
    \label{eq:weight}
    w_i=\exp \left(-\sum_{j=1}^{i-1} \boldsymbol{\sigma}\left(\mathbf{x}_j\right)\right) \left(1-\exp \left(-\boldsymbol{\sigma}\left(\mathbf{x}_i\right)\right)\right)
\end{equation}

\noindent where $\sigma(\mathbf{x}_i) = \frac{1}{\alpha} \cdot \operatorname{Sigmoid} \left(\frac{-\phi_{\boldsymbol{g}}(\mathbf{x}_i)}{\alpha}\right) $
is the 3D volumetric density that can be converted from the SDF $\Phi_g(\mathbf{x}_i)$ \cite{stylesdf}, $\alpha$ is a learnable parameter which controls the sharpness of the model.  This method of conversion through density, compared to direct conversion \cite{coslam, voxfusion} and surface-based conversion \cite{neus, goslam}, offers better interpretability, aligning closely with the original volumetric rendering in NeRF \cite{mildenhall2021nerf}. Moreover, we leverage this representation to derive our definition of uncertainty, which will be discussed in the following section.

\subsection{Uncertainty Modeling}
\label{sect:uncertaintyModel}

Our primary objective is to derive an uncertainty measure that can indicate the quality of the color and depth images, allowing us to reweight the loss functions during tracking and mapping. However, to our knowledge, no NeRF-based dense SLAM system has yet addressed
predictive uncertainty, which reflects the model’s confidence explicitly in its predictions for each view.

Specifically, inspired by the vanilla NeRF formulation \cite{mildenhall2021nerf} (Eq. 3), we utilize the \textbf{termination probability} concept from the volume rendering equation.
\vspace{-2mm}
\begin{equation}
    \label{eq:activeUncertainty}
    w_i=\overbrace{\underbrace{\exp \left(-\sum_{j=1}^{i-1} \boldsymbol{\sigma}\left(\mathbf{x}_j)\right)\right)}_{\text {transmittance }T_i} \underbrace{\left(1-\exp \left(-\boldsymbol{\sigma}\left(\mathbf{x}_i\right)\right)\right)}_{\text {occupancy }o_i}}^{\text{termination probability at sample }i} = T_i \cdot o_i
\end{equation}
\vspace{-3mm}

\noindent where $T_i$ describes \textit{transmittance} at sample point $ t_i $  along the ray from $t_0$ to $t_{i-1}$ without hitting any other particle, \textit{occupancy }$o_i$ represents the probability that the ray collides with a particle at position $t_i$ independently of the previously light path. The product of the two $ w_i = T_i \cdot o_i$  represents the termination probability, i.e. the probability that the light can reach the spatial location  $t_i$.

We define the accumulated termination probability of $N$ sampling points along a current sampling ray $r$ as
\begin{equation}
    p(r)=\sum_{i=1}^N w_i = 1-\exp \left(-\sum_{i=1}^N \sigma\left(\mathbf{x}_i\right)\right)
    \label{eq:terminationProbalility}
\end{equation}

\noindent $p(r) = 1$ ideally when the rendering is perfect (camera tracking is accurate and the region has been already observed before). Conversely, in never unobserved regions the NeRF model will estimate a low termination probability $p(r)\approx 0$ along the current ray $ r $. The value is bounded by $(0,1)$. We validate this in our experiments and visualize the termination probability in \cref{fig:termination_probability} (d), and the mathematical proof is included in the supplementary material Sec. A.2.

\begin{figure}[tb]
  \centering
  \includegraphics[height=8.5cm]{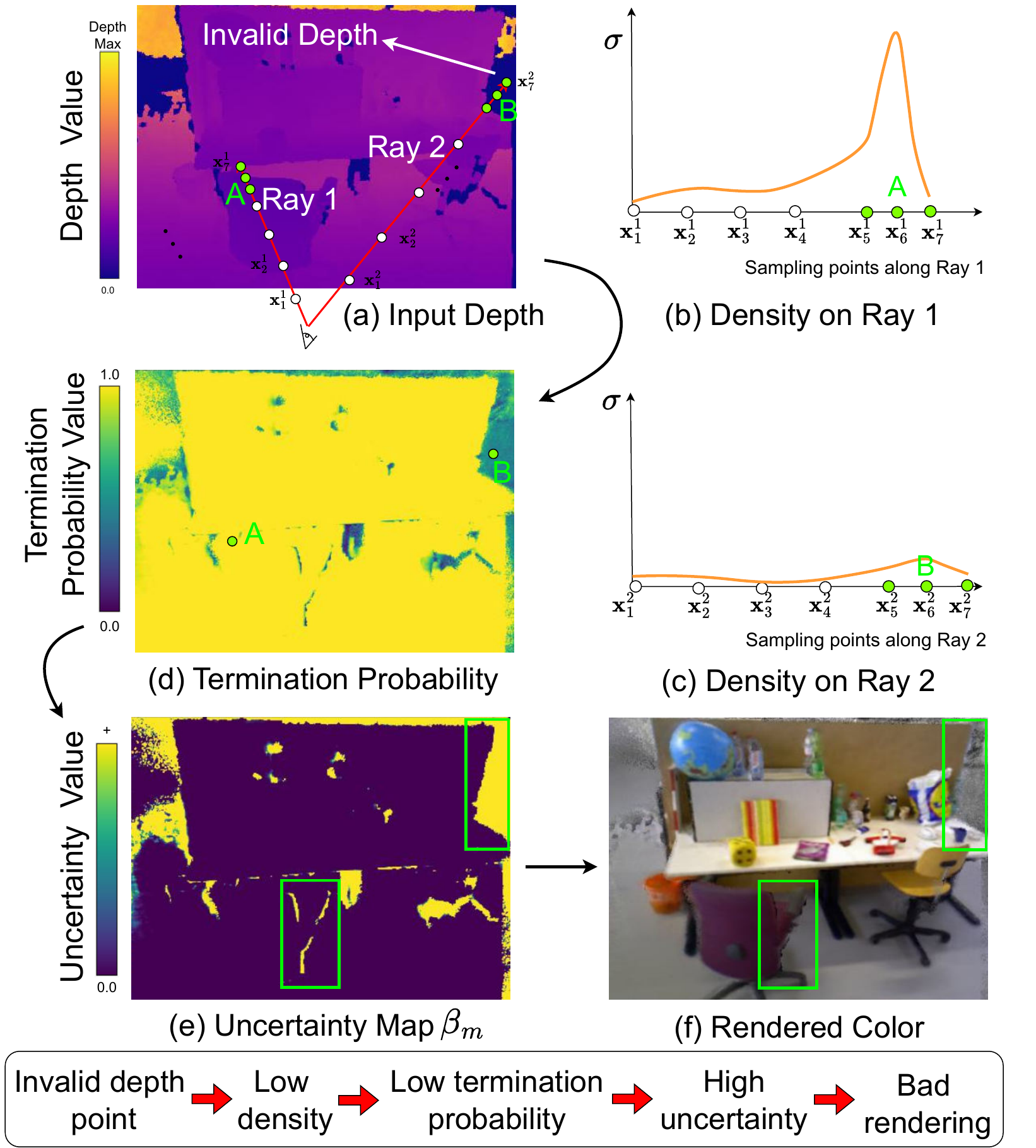}
  
  \caption{\textbf{Termination Probability and Uncertainty.}  This figure illustrates the termination probability and uncertainty during ray sampling. For pixel A with valid depth (sampling by Ray 1), the sampling density is high along this ray, leading to a high termination probability and lower uncertainty. In contrast, for pixel B with invalid depth (sampling by Ray 2), the sampling density is low along this ray, resulting in a lower termination probability and higher uncertainty, as seen in the uncertainty map (e). This leads to degraded rendering quality in regions with high uncertainty, as shown in (f). Back-projected points A and B correspond to the surfaces of the hit objects in 3D space. For point B with invalid depth, we can estimate an approximate depth value based on the model in its current state.}
  \label{fig:termination_probability}
 \vspace{-15pt}
\end{figure}

In \cite{sunderhauf2023density} S{\"u}nderhauf et al. define uncertainty based on deep ensembles. However, full deep ensembles require training multiple models with different initializations, and are unsuitable for real-time systems like SLAM due to the high computational cost of maintaining several models. 
For a given image with an estimated pose, a pixel with index  $m$ is associated to a corresponding ray $r_m$. Inspired by \cite{sunderhauf2023density}, based on the probability $p(r_m)$, we defined the pixel-level predictive uncertainty as
\begin{equation}
    \beta_m = \left(1-p(r_m)\right)^2
    \label{eq:pixeleveluncertainty}
\end{equation}
As shown in \cref{fig:termination_probability}(a), pixel B with invalid depth, we can only estimate an approximate depth value based on the model in its current state. Using this estimated depth for ray sampling results in a rendering with low accumulated termination probability in \cref{fig:termination_probability}(d), indicating higher uncertainty as seen in \cref{fig:termination_probability}(e) the uncertainty map.

For a rendered image associated with $M$ sampled rays, we introduce a novel image-level predictive uncertainty $\beta$ defined as
\vspace{-2mm}
\begin{equation}
    \label{eq:modelUncertainty}
    \beta= \frac{1}{M} \sum_{m=1}^M \beta_m
\end{equation}

\noindent This image-level uncertainty $\beta$ indicates the model's confidence in its current position estimate. A low $\beta$ value suggests that the model is familiar with the area because of the accurate estimated camera position and sufficient sampling rays. Conversely, a high $\beta$ value indicates that the model is less familiar with the area, suggesting that it should be more cautious and attentive in this region.

This predictive uncertainty, reflecting the model's knowledge limitations on the current camera pose, can be reduced by gathering more data, such as by taking data slowly to avoid drastic changes in motion state. How to use the defined uncertainty in the loss function and keyframe selection will be discussed in \cref{sect:lossFunction} and \cref{sect:dynamicBA}.

We also compared our model-free uncertainty approach with a learnable uncertainty model, based on Gaussian assumptions, as in BayesRays \cite{bayesrays}. Our experiments show that this idea not only brings undesirable increased model complexity, making the model much slower, but also leads to poorer results in terms of reconstruction quality. Details can be found in the supplementary material Sec. B.3.

\subsection{Uncertainty-guided Loss Function}
\label{sect:lossFunction}
Our mapping and tracking processes are carried out by minimizing our objective functions with respect to the network parameters $\theta$ and the camera parameters $\{R_i | t_i\}$ as \cite{coslam}. We hypothesize that pixels with invalid depth or motion blur, caused by sensor issues or sudden motion changes, should exhibit high uncertainty, while well-observed regions should display low uncertainty. This premise enables us to effectively incorporate predicted uncertainty into the objective function, with the goal of progressively filtering out outliers to enhance localization accuracy and rendering quality. Inspired by the definition of SSIM loss in NeRF on-the-go \cite{nerfOn-the-go} and the masked uncertainty learning in DebSDF \cite{debsdf}, we define pixel-level binary confidence function as 

\vspace{-5mm}
\begin{equation}
CF_m= \mathds{1}\left(1 - \beta_m \right) =  \begin{cases}1 & \text { if } \beta_m \leq \beta_{unc_m} \\ 0 & \text { if } \beta_m>\beta_{unc_m}\end{cases}
\label{eq:binary_confidence}
\end{equation} 
where $\beta_{unc_m}$ is a threshold for pixel-level uncertainty.

Near the surface we set hyperparameter truncation distance $\tau_{\text {trunc}}$ and approximate the ground truth SDF of sampling point $\mathbf{x}_i$ by $b\left(\mathbf{x}_i\right)=D_m-D_{m, i}^{r a y}$, where $D_m$ is current ray depth,  $D_{m, i}^{r a y}$ is the distance from camera center $
\text { w.r.t.~}$ sampling point. The points that lie within the truncation distance $[- \tau_{\text {trunc }}, \tau_{\text {trunc }}]$,  $i.e. |b\left(\mathbf{x}_i\right)|< \tau_{\text {trunc}}$ form the set $X^{tr}$. 

The loss associated to the points belonging to $X^{tr}$ is
\vspace{-5pt}
\begin{equation}
    \mathcal{L}^{tr}(X^{t r}) =\frac{1}{M^*} \sum_{m=1}^M \frac{CF_m}{\left|X^{t r}\right|}
 \sum_{\mathbf{x}_i \in X^{t r}}\left(\Phi_g(\mathbf{x}_i) \tau_{\text {trunc }} - b\left(\mathbf{x}_i\right)\right)^2 
\end{equation}
where $M$ is the number of sampled points, $M^*$ is the number of valid sampled points after reweighting by \cref{eq:binary_confidence}.

We further refine the set of sampling points inside the truncation distance in two subgroups. Assuming accurate valid depth ground truth, we assign greater weights to sample points at the \textit{center} (closer to the surface) $X_c^{t r}=\{\mathbf{x}_i \mid \left.\  |b\left(\mathbf{x}_i\right)|  \leq 0.4 \tau_{\text {trunc}}\right\}$ to accelerate convergence and achieve more accurate geometry, while points at the \textit{tail} of the truncation region constitute $X_t^{t r}$, and associate different losses to these two groups as follows:
\begin{equation}
    \mathcal{L}_{c}^{tr}=\mathcal{L}^{tr}(X_c^{t r})  \quad \text{and} \quad  \mathcal{L}_{t}^{tr}=\mathcal{L}^{tr}(X_t^{t r})
\end{equation}

Considering the points outside the truncation distance as the free space set $X^{fs}$, which are far from the surface $|b\left(\mathbf{x}_i\right)|> \tau_{\text {trunc}}$. In this area the loss function encourages $\Phi_g(\mathbf{x}_i)$ to have the value equal to one as
\vspace{-3mm}
\begin{equation}
    \mathcal{L}^{fs}=\frac{1}{M^*} \sum_{m=1}^M \frac{CF_m}{\left|X^{fs}\right|}\sum_{\mathbf{x}_i\in X^{f s}}\left(\Phi_g(\mathbf{x}_i)-1\right)^2 
\end{equation}
\vspace{-5pt}

The color and depth losses are defined as follows:
\vspace{-6pt}
\begin{equation}
    \mathcal{L}_{r g b}^{track}=\frac{1}{M^*} \sum_{m=1}^M \left(C[u, v] - \hat{\boldsymbol{c}}_m\right)^2 \cdot CF_m
\end{equation}
\vspace{-6pt}
\begin{equation}
    \mathcal{L}_{r g b}^{map}=\frac{1}{M} \sum_{m=1}^M \left(C[u, v] - \hat{\boldsymbol{c}}_m\right)^2
    \label{eq:mapping_loss}
\end{equation}
\vspace{-6pt}
\begin{equation}
    \mathcal{L}_{d e p}=\frac{1}{M^*} \sum_{m=1}^M \left(D[u, v] - \hat{\boldsymbol{d}}_m\right)^2 \cdot CF_m
\end{equation}
where $C[u, v]$ and $D[u, v]$ are the ground-truth values for color and depth respectively. Note the reweighting confidence function $CF_m$ is not applied to color loss in the mapping process. 

\noindent\textbf{Tracking Loss Function.}  The loss function for the tracking process is achieved by the following weighting scheme: 
\begin{equation}
\mathcal{L}_t = \lambda_{rgb} \mathcal{L}_{rgb}^{track} + \lambda_{dep} \mathcal{L}_{dep} + \mathcal{L}_{sdf}
\end{equation}

\noindent where $\mathcal{L}_{sdf} = \lambda_{c}^{tr} \mathcal{L}_{c}^{tr} + \lambda_{t}^{tr} \mathcal{L}_{t}^{tr} + \lambda_{{fs}} \mathcal{L}_{{fs}}$.

During tracking, the scene representation remains unchanged and only the camera pose is optimized (as shown by the magenta dashed line in \cref{fig:activeSLAMPipeline}). $CF_m$ helps us select the most confidently estimated data for optimal optimization. If certain pixels are already predicted incorrectly, continuing to assign them a high weight is not beneficial. Therefore, when applying the tracking loss function, it is crucial to focus on pixels that are correctly estimated with high confidence. This means that the loss for pixels which are misestimated with high uncertainty can be neglected.

\vspace{2mm}
\noindent\textbf{Mapping Loss Function.} 
The total loss function for mapping loss is defined as:
\begin{equation}
\mathcal{L}_m = \lambda_{rgb} \mathcal{L}_{rgb}^{map} + \lambda_{dep} \mathcal{L}_{dep} + \mathcal{L}_{sdf}
\end{equation}

Unlike tracking, the mapping process relies more on RGB information to compensate for invalid depth, requiring a distinct treatment of $\mathcal{L}_{rgb}^{tracking}$ and $\mathcal{L}_{rgb}^{map}$. Additionally, since scene representation is optimized only during mapping, we do not reweight $\mathcal{L}_{rgb}^{map}$ with the confidence function $CF_m$ in \cref{eq:mapping_loss}.

\subsection{Strategic Bundle Adjustment}
\label{sect:dynamicBA}
\begin{figure}[tb]
  \centering
  \includegraphics[height=3cm]{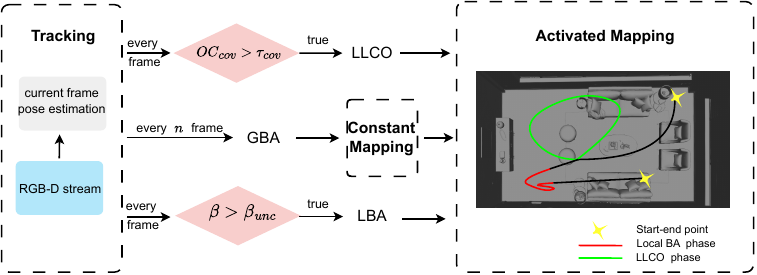}
  \caption{\textbf{Strategic BA.} While the tracking process is performed at every frame, we perform a constant mapping with global bundle adjustment (GBA) at a fixed frequency. Thus, the pose and map are optimized using all keyframes from the start to the end of the frame sequence. If an outlier frame is detected based on its uncertainty, a local bundle adjustment (LBA) is performed, as shown in \textcolor{red}{red}. If a loop closure is detected, a local loop closure optimization (LLCO) is performed, as shown in \textcolor{green}{green} in the figure.}
  \label{fig:BA_selection}
  \vspace{-6mm}
\end{figure}

\begin{figure}[htbp]
  \centering
  \includegraphics[height=3cm]{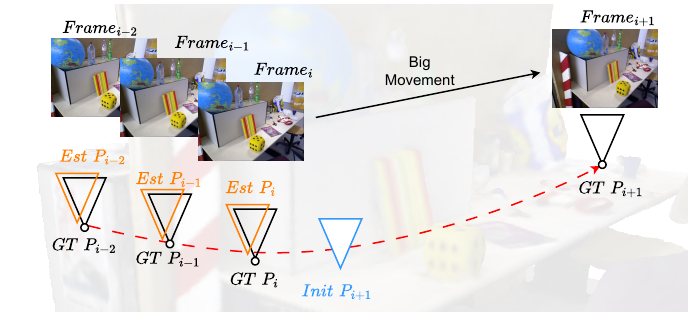}
  \caption{\textbf{Activated additional local BA}. From position $P_i$ to $P_{i+1}$, sudden large movements lead to difficulties in pose estimation and increased uncertainty due to unseen areas. The initialization of $Init \ P_{i+1}$ based on the constant speed assumption is hard to optimize. Therefore, besides constant global BA, we activate additional local BA based on image-level uncertainty to optimize local information. This simulates slowing down the movement. Its effectiveness can be found in \cref{fig:tumrgbd_mesh} and \cref{tab:impact_BA}.
}
  \label{fig:cap_local}
  \vspace{-3mm}
\end{figure}

In bundle adjustment (BA), keyframes are selected first, followed by joint optimization of camera poses and scenes. Traditional dense SLAM techniques require storing keyframe images for dense pixel-level loss calculation. Recent NeRF-based SLAM methods like iMap \cite{imap} and Nice-SLAM \cite{nicerslam} use local BA, selecting a small fraction of keyframes and points through a sliding window. In \cite{coslam, birn-slam}, global BA optimizes all keyframes. However, none of these NeRF-based SLAM methods incorporate uncertainty management in keyframe selection or BA. Performing mapping process every $n$ frames is unreasonable due to the random motion states and varying quality of depth and color images, which provide different information to the scene representation. Any misestimation (e.g., outlier pose) will have a global impact and might cause false reconstruction. Therefore, corrective and remedial strategies are needed. 
To better balance efficiency and accuracy, we propose an uncertainty-guided local-to-global bundle adjustment, as depicted in \cref{fig:BA_selection}. Tracking operations are executed for every frame, while mapping with global BA occurs every $n$ frames constantly. In order to capture local information, our Uni-SLAM system can activate \textbf{additional mapping processes} with local BA based on image-level uncertainty $\beta$ if $\beta > \beta_{unc}$, where $\beta_{unc}$ is the threshold for image-level uncertainty. In local BA, we use only keyframes that visually overlap with the current frame, mitigating the impact of outlier frames. \cref{fig:cap_local} illustrates this necessity.

For local BA keyframe selection, we first initialize spatial sample points in 3D space using the current frame's camera pose. These points are then back-projected onto previous keyframes to check how many fall within image boundaries, determining overlap. Prioritizing local over global information, this method enables efficient local map updates with a limited number of $M$ sample points and informs our co-visibility check.
\cref{eq:covibility} defines the overlapping coefficient of co-visibility $OC_{cov}(i, c)$ between $i$-keyframe $I_i$ and current frame $ I_c $,  $I_i \in \text{\textit{Keyframe Database}} \left\{ I_1, I_2, \ldots, I_n \right\}  $

\begin{equation}
    OC_{cov}(i, c)=\frac{\left|I_i \cap I_c\right|}{\left|I_c\right|}
\label{eq:covibility}
\end{equation}

At the end of the tracking process for every frame, we calculate the co-visibility with negligible computational overhead. If the co-visibility is larger than threshold $\tau_{cov}$ (set at 0.95), it indicates a loop closure. In this case, the \textbf{additional mapping process} with local loop closure optimization (LLCO) is performed immediately. This process optimizes only the keyframes from the current frame to the loop closure point, as shown in \textcolor{green}{green} circle in \cref{fig:BA_selection}. This approach enables efficient use of $M$ sample points and improves system stability.

\vspace{-3mm}
\section{Experiments and Results}
\label{sec:experiment}

\subsection{Experimental Setup}
\label{sec:experimentalSetup}

\noindent\textbf{Datasets.} We evaluate Uni-SLAM using diverse benchmarks, including the synthetic Replica dataset\cite{replica} with 8 high-quality indoor scene reconstructions, as well as the realistic ScanNet\cite{scannet} and TUM RGB-D datasets\cite{tumrgbd}.

\noindent\textbf{Metrics.} We assess the quality of our reconstruction from multiple perspectives. For tracking accuracy, we adopt \emph{ATE RMSE [cm]} \cite{ateRMSE}. We analyze the reconstruction quality using 3D and 2D metrics. For 3D metrics, the meshes produced by marchingcubes \cite{marchingcube} are evaluated by \emph{Depth L1 [cm], Accuracy [cm], Reconstruction completion [cm]},  and \emph{Completion ratio [1cm ]\%}. Those meshes are culled following \cite{neuralrgbd} before evaluation. For 2D rendering, we provide the peak signal-to-noise ratio (PSNR), SSIM \cite{ssim}, and LPIPS \cite{lpips}. The rendering metrics are evaluated every 5 frames on full-resolution images.

\noindent\textbf{Baselines and Implementation.}
 We primarily compare our method to existing state-of-the-art dense implicit RGB-D SLAM systems such as Nice-SLAM\cite{niceslam}, Co-SLAM\cite{coslam}, ESLAM\cite{eslam}, and BSLAM\cite{birn-slam}. For BSLAM we produce results with their novel proposed hybrid model. We reproduce their results using the open-source code and report the middle value after 5 runs. The results of iMAP$^*$ \cite{imap} are adopted from Nice-SLAM. For a fair comparison, we extract mesh at $1cm$ resolution. In our pipeline implementation, we set the hash grid level to 16 for both geometry and appearance grids. We randomly select 4,000 sampling points for the mapping process and 2,000 for the tracking process. The truncation distance is set to 6 cm. Additional details can be found in Supp. Sec. A.1.

\begin{figure*}[t]
\captionsetup{skip=2pt}
  \centering
  \captionsetup{skip=-2pt}
  \includegraphics[height=6 cm]{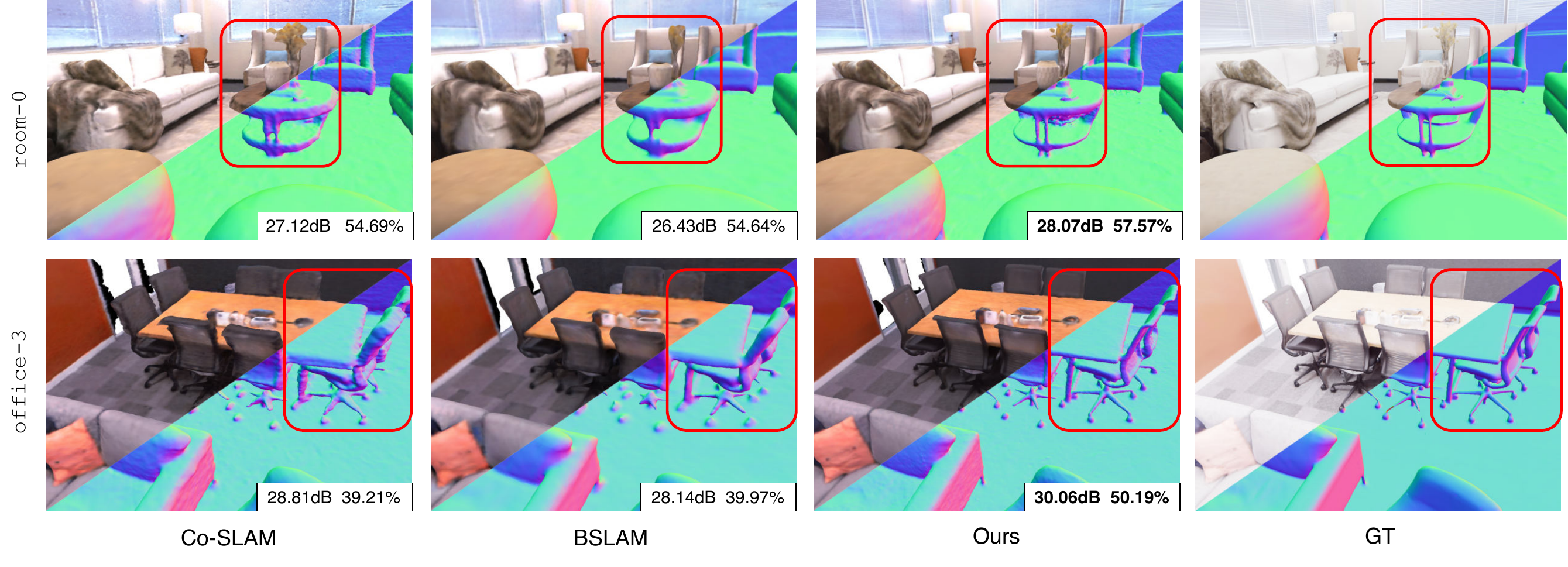}
  \caption{\textbf{Mesh Evaluation on Replica \cite{replica}.} Our method outstands with its thin geometry details and higher texture fidelity compared to Co-SLAM\cite{coslam} and BSLAM\cite{birn-slam}. For example, the table and vase in \texttt{room-0}; the thinner office desk, chair backrest, and detailed reconstructed chair legs in \texttt{office-3}. In the lower right corner, we note rendering quality in PSNR$[dB]\uparrow$ and geometric evaluation in completion ratio $[<1 cm\%]\uparrow$.  Please zoom-in for more details.}
  \label{fig:replica_mesh}
\end{figure*}

\vspace{-3mm}
\begin{figure*}[htbp]
  \centering
  \captionsetup{skip=-1pt}
  \includegraphics[height=4 cm]{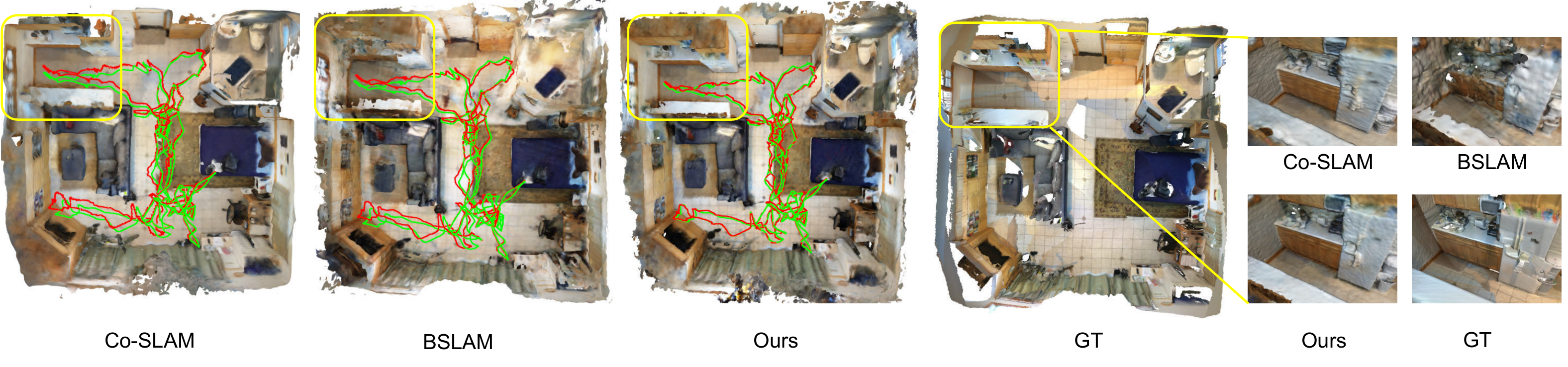}
  \caption{\textbf{Mesh Evaluation on ScanNet \cite{scannet}.} The estimated pose is shown in \textcolor{red}{red}, and the ground truth camera pose is shown in \textcolor{green}{green}. Our method stands out with its more accurate trajectory and higher quality reconstruction, such as the corners of the kitchen.}
  \label{fig:scannet_mesh}
  \vspace{-5mm}
\end{figure*}

\begin{figure}[htbp]
  \centering
  \captionsetup{skip=-5pt}
  \includegraphics[height=6cm]{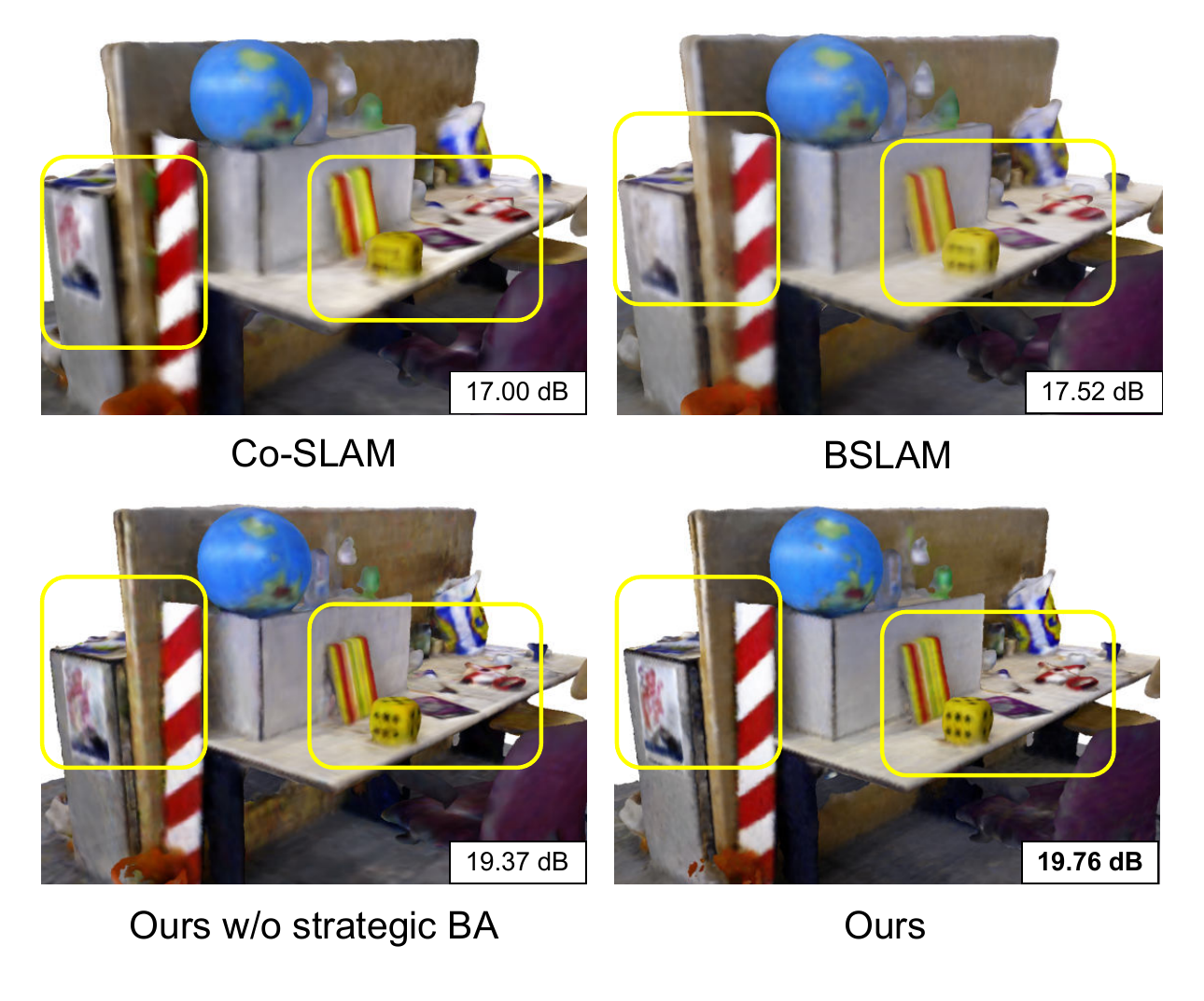}
  \caption{\textbf{Mesh Evaluation on TUM RGB-D \cite{tumrgbd}.} Our method stands out with its geometry details and higher texture fidelity. Without strategic BA (only with global BA), the performance can be suboptimal due to missing local information.}
  \label{fig:tumrgbd_mesh}
\vspace{-7mm}
\end{figure}

\begin{table}[htbp]
\scriptsize
\centering
\begin{tabular}{l | c @{\hspace{3pt}}  c @{\hspace{3pt}} c @{\hspace{3pt}} c @{\hspace{3pt}} c @{\hspace{3pt}} c @{\hspace{3pt}} c @{\hspace{3pt}}c | @{\hspace{3pt}} l}
\hline 
Method    & Rm 0  & Rm 1 & Rm 2 & Off 0 & Off 1 & Off 2 & Off 3 & Off 4& Ave. \\
\hline 
iMAP$^*$  \cite{imap}      & 5.23 & 3.09 & 2.58 & 2.4 & 1.17 & 5.67 & 5.08 & 2.23 & 3.24 \\

Nice-SLAM \cite{niceslam}  & 0.97 & 1.31 & 1.07 & 0.88 & 1.00 & 1.06 & 1.10 & 1.13 & 1.06\\
MIPS-Fusion \cite{mipsFusion} & 1.10 & 1.20 & 1.10 & 0.70 & 0.80 & 1.30 & 2020 & 1.10 & 1.19\\
Co-SLAM \cite{coslam}    & 0.66 & 2.25 & 1.07 & 0.65 & 0.53 & 2.12 & 1.32 & 0.85 & 1.18 \\

ESLAM \cite{eslam}      & 0.69 & 0.70 & 0.52 & 0.57 & 0.55 & 0.58 & 0.72 & 0.63 & 0.63 \\

BSLAM \cite{birn-slam}  & 0.71 & 0.88& 1.5 & 0.61 & 0.49 & 2.14 & 1.63 & 1.66 & 1.19 \\


Ours     & \textbf{0.49} & \textbf{0.48} & \textbf{0.40} & \textbf{0.37} & \textbf{0.36} & \textbf{0.48} & \textbf{0.56} &\textbf{0.44}  &\textbf{0.45} \\
\hline
\end{tabular}
\caption{\textbf{Tracking performance on Replica \cite{replica}}(RMSE ↓ [cm]).}
\vspace{-3mm}
\label{tab:trackingReplica}
\end{table}

\begin{table}[htbp]
\tiny
\centering
\begin{tabular}{l | c@{\hspace{2pt}} c@{\hspace{2pt}} c@{\hspace{2pt}} c  | c@{\hspace{2pt}} c@{\hspace{2pt}} c}
\hline 
\multirow{2}{*}{Method} & \multicolumn{4}{c|}{Reconstruction $[cm]$} & \multicolumn{3}{c}{Rendering} \\
& \tiny Depth L1$\downarrow$ & \tiny Acc.$\downarrow$ & \tiny Comp.$\downarrow$ & \tiny Comp. Ratio $[\%] \uparrow$  & \tiny PSNR$[dB]\uparrow$ & \tiny SSIM$\uparrow$ & \tiny LPIPS$\downarrow$  \\
\hline 
iMAP*\cite{imap}            & 8.23 & 7.16 & 5.83 & 20.33 & 17.32 & 0.6535 & 0.3425\\
Nice-SLAM\cite{niceslam}    & 3.18 & 1.90 & 1.53 & 36.93 & 24.42 & 0.8091  & 0.2335 \\Co-SLAM\cite{coslam}        & 2.15 & 1.16 & 1.12 & 55.94 & 30.27 & 0.9396 & 0.2468\\
ESLAM\cite{eslam}           & 1.18 & 0.97 & 1.05 & 63.99 & 30.19 & 0.9421 & 0.2433 \\
BSLAM\cite{birn-slam}       & 2.52 & 1.12 & 1.10 & 57.18 & 29.55 & 0.9335 & 0.2361 \\
Ours                        & \textbf{0.89} & \textbf{0.92} & \textbf{0.92} & \textbf{66.86} & \textbf{31.62} & \textbf{0.9584} & \textbf{0.1853}\\
\hline
\end{tabular}
\caption{\textbf{Reconstruction and Rendering Performance on Replica}\cite{replica}\textbf{.} To reflect the ability to reconstruct geometric details, we report completion ratio $[<1 cm\%]$. For the details of the evaluations for each scene, refer to the supplementary material.}
\label{tab:replica_recon_rendering}
\end{table}

\begin{table}[t]
\scriptsize
\centering
\resizebox{8.5cm}{!}{\begin{tabular}{l l  | c @{\hspace{5pt}}  c @{\hspace{5pt}} c @{\hspace{5pt}} c @{\hspace{5pt}} c @{\hspace{5pt}} c | @{\hspace{5pt}}c}
\hline 
Method &   & Sc.00 & Sc.59 & Sc.106 & Sc.169 & Sc181 & Sc.207 & Ave. \\

\hline 
\multirow{1}{*}{iMAP*\cite{imap}}& \multirow{1}{*}{ICCV 21}  & 42.7 & 17.8 & 15.0 & 39.1 & 24.7 & 20.1 & 26.6 \\

\multirow{1}{*}{Nice-SLAM\cite{niceslam}}& \multirow{1}{*}{CVPR 22}  & 12.0 & 14.0 & 7.9 & 10.9 & 13.4 & 6.2 & 10.7 \\

\multirow{1}{*}{MIPS-Fusion\cite{mipsFusion}}& \multirow{1}{*}{SA 23}  & 7.9 & 10.7 & 9.7 & 9.7 & 14.2 & 7.8 & 10.0 \\

\multirow{1}{*}{ESLAM\cite{eslam}}& \multirow{1}{*}{CVPR 23}  & 7.3 & 8.5 & 7.5 & 6.5 & \textbf{9.0} & 5.7 & 7.4 \\
 
\multirow{1}{*}{Co-SLAM\cite{coslam}}& \multirow{1}{*}{CVPR 23}  & 7.2 & 12.3 & 9.6 & 6.6 & 13.4 & 7.1 & 9.4 \\

\multirow{1}{*}{BSLAM\cite{birn-slam}}& \multirow{1}{*}{CVPR 24}  & 7.29 & 12.2  & 9.0  & 8.8  & 13.4 &6.65 & 9.56 \\

\multirow{1}{*}{Ours} & \multirow{1}{*}{}  & \textbf{6.12} & \textbf{7.77} & \textbf{7.41} & \textbf{5.82} & 9.77 & \textbf{5.21} & \textbf{7.01} \\
\hline
\end{tabular}}
\caption{\textbf{Tracking Performance on ScanNet}\cite{scannet}(RMSE [cm]) \textbf{.} On average, our method achieved the best results.}
\label{tab:trackingScanNet}
\vspace{-5mm}
\end{table}

\vspace{-2mm}
\subsection{Qualitative and Quantitative Evaluation}
\label{quantitativeEvaluation}

\noindent\textbf{Reconstruction \& Rendering.}
\cref{{fig:replica_mesh}} compares the mesh reconstructions of Co-SLAM\cite{coslam}, BSLAM\cite{birn-slam} and ours to ground truth mesh on Replica. Our method can achieve more accurate thin geometric details and high-fidelity colors, such as captured chair legs and thin tables. Quantitatively, \cref{tab:replica_recon_rendering} compares reconstruction and rendering performance on the Replica dataset and shows best among 3D metrics and 2D metrics, beating all implicit dense SLAM. In \cref{fig:scannet_mesh} we show that our method can achieve more accurate localization and finer realistic details on ScanNet.  We attribute this to our sufficient model capability and online uncertainty-aware activated additional mapping process, which can capture more details locally. The reconstructed mesh on TUM RGB-D is shown in \cref{fig:tumrgbd_mesh}. The results show that our reconstruction quality benefits from strategic BA.

\noindent\textbf{Tracking.}
\cref{tab:trackingReplica} compares our methods to state-of-the-art implicit dense RGB-D neural SLAM system on 8 scenes of Replica datasets\cite{replica} in tracking performance.  We outperform on all scenes and achieve an average improvement of $62\%$, $29\%$ and $62\%$ on RMSE over Co-SLAM, and ESLAM and BSLAM respectively. The tracking performance on ScanNet and TUM RGB-D is shown in \cref{tab:trackingScanNet} and \cref{tab:trackingTUMrgbd} respectively. We primarily attribute this to the uncertainty reweighted loss function, where only the most reliable information is emphasized. Although classic methods are still showing state-of-the-art accurate tracking on TUM RGB-D, our method outperforms neural methods on average and bridges the gap between those two categories.

\begin{table}[htbp]
\scriptsize
\centering
\resizebox{5cm}{!}{
\begin{tabular}{l | l | c @{\hspace{5pt}} c @{\hspace{5pt}} c @{\hspace{5pt}} c}
    \hline 
    \centering
    \multirow{2}{*}{} & \multirow{2}{*}{Method} & fr1/ & fr2/ & fr3/ & \multirow{2}{*}{Ave.} \\
                         &     & desk & xyz  & office & \\
    \hline 
    \multirow{6}{*}{NeRF-Based} & iMAP$^*$ \cite{imap}            & 5.15 & 2.39 & 5.76 & 4.43 \\
    & Nice-SLAM \cite{niceslam}       & 5.00 & 3.17 & 5.05 & 4.41 \\
    & MIPS-Fusion \cite{mipsFusion}   & 3.00 & 1.40 & 4.6  & 3.0 \\
    &  Co-SLAM \cite{coslam}           & 3.05 & 1.88 & 2.85 & 2.59 \\
    & ESLAM \cite{eslam}              & 2.54 & \textbf{1.13} & 2.75 & 2.14 \\
    & BSLAM \cite{birn-slam}          & 2.87 & 1.38 & 2.95 & 2.39 \\
    & Ours                            & \textbf{2.37} & 1.17 & \textbf{2.62} & \textbf{2.05} \\
    \hline
    \multirow{3}{*}{Classic} & ORB-SLAM2 \cite{orbslam}         & \textbf{1.6} & \textbf{0.4} & \textbf{1.0} & \textbf{1.0} \\
    & BundleFusion \cite{bundlefusion} & 1.6 & 1.1 & 2.2 & 1.63 \\
    & BAD-SLAM \cite{badslam}          & 1.7 & 1.1 & 1.7 & 1.5 \\
    \hline
\end{tabular}
}
\caption{\textbf{Tracking Performance on TUM RGB-D}\cite{tumrgbd} (RMSE [cm]) \textbf{.}}
\label{tab:trackingTUMrgbd}
\vspace{-5mm}
\end{table}

\subsection{Analysis on Design Choices}
\label{ablation}

\noindent \textbf{Runtime and Memory Analysis } 
In \cref{tab:runtimeMemory}, we compare runtime and memory usage, benchmarking all methods on NVIDIA GeForce RTX 4090 GPU using \texttt{room0} of Replica \cite{replica}. We report tracking and mapping times per iteration and compare iteration steps to show convergence speed. Our model achieves real-time performance on par with SOTA results at speeds exceeding 8 FPS.

\begin{table}[h]
\vspace{-2mm}
\tiny
\centering
\begin{tabular}{ l | c @{\hspace{5pt}}  c @{\hspace{5pt}} c @{\hspace{5pt}} c @{\hspace{5pt}} c @{\hspace{5pt}} }
\hline 
  \multirow{2}{*}{Method}    & Tracking                 & Mapping           & \multirow{2}{*}{FPS$\uparrow$} & Time & \multirow{2}{*}{Params.$\downarrow$} \\ 
                              & [ms x it.] $\downarrow$     & [ms x it.] $\downarrow$ &  & Mins$\downarrow$                       \\        
\hline 

                        Nice-SLAM \cite{niceslam}      & 6.5 x 10 & 29.3x60  & 1.8   & 18.51 & 12.13M \\

                        Co-SLAM \cite{coslam}          & \textbf{4.6 x 10} & \textbf{6.6 x 10}   & \textbf{9.07}  & \textbf{3.67}  & \textbf{1.72M} \\

                        ESLAM \cite{eslam}             & 7.9 x 8  & 18.8 x 15 & 5.55  & 6.01  & \underline{6.78M} \\

                        BSLAM \cite{pointslam}         & 11 x 20  & 15 x 20    & 1.66   & 20.3  & 17.38M \\

                        Ours                           & \underline{7.0 x 8} & \underline{8.1 x 13}  & \underline{8.37}   & \underline{4.02}  & 12.69M \\
                        
\hline 
\end{tabular}
\caption{Runtime and Memory Usage Comparison.}
\label{tab:runtimeMemory_main_main}
\vspace{-2mm}
\end{table}

\noindent\textbf{Ablation of Model Design.}
We encoded geometry and appearance using different structures and validated our design choices on the Replica dataset \cite{replica}, as shown in \cref{tab:ablation_model_design_main}. By ablating various combinations of hash grids \cite{instantngp} and tri-planes \cite{chan2022efficient}, we found that using two hash grids without a third learnable uncertainty grid (h-h-n) produced the best results. Introducing a third learnable uncertainty grid (h-h-u) under the Gaussian assumption made training and convergence more complex. Further details can be found in Supplementary Sec. B.3.

\begin{table}[htbp]
\tiny
\vspace{-2mm}
\centering
\begin{tabular}{l | c@{\hspace{2pt}} c@{\hspace{2pt}} c@{\hspace{2pt}} c  | c@{\hspace{2pt}} c@{\hspace{2pt}} c}
\hline 
\multirow{2}{*}{Method} & \multicolumn{4}{c|}{Reconstruction $[cm]$} & \multicolumn{3}{c}{Rendering/Tracking/Time} \\
& \tiny Depth L1$\downarrow$ & \tiny Acc.$\downarrow$ & \tiny Comp.$\downarrow$ & \tiny Comp. Ratio $[\%] \uparrow$  & \tiny PSNR$[dB]\uparrow$ & \tiny RMSE$[cm]\downarrow$ & \tiny Mins$\downarrow$  \\
\hline 
h-h-u                        & 3.75 & 1.79 & 1.65 & 31.52 & 27.33 & 1.51 & 6.53 \\
h-t-n                        & 0.93 & 1.01 & 1.15  & 64.69 & 30.98 & 0.47 & 4.79 \\
t-h-n                        & 0.97 & 1.17 & 1.09  & 63.82 & 31.32 & 0.50 & 4.65 \\
Ours(h-h-n)                  & \textbf{0.89} & \textbf{0.92} & \textbf{0.92} & \textbf{66.86} & \textbf{31.62} & \textbf{0.45} & \textbf{3.97}\\
\hline
\end{tabular}
\caption{Ablation of model design.}
\label{tab:ablation_model_design_main}
\vspace{-5mm}
\end{table}

\begin{figure}[htbp]
  \vspace{-3mm}
  \centering
  \captionsetup{skip=-2pt}
  
  \begin{subfigure}[c]{0.01\linewidth} 
    \captionsetup{skip=-2pt}
    \tiny
    \begin{tabular}{l | c@{\hspace{3pt}} c@{\hspace{1pt}} c@{\hspace{1pt}} | c@{\hspace{1pt}} c@{\hspace{1pt}} c@{\hspace{1pt}}}
    \hline 
    \multirow{2}{*}{Method} & \multicolumn{3}{c|}{Reweighting Term} & \multicolumn{3}{c}{Tracking/Rendering} \\
      & \tiny SDF & \tiny Depth & \tiny Color & \hspace{-3pt} \tiny RMSE [cm] $\downarrow$ & \tiny PSNR [dB] $\uparrow$  & \tiny Depth L1[m]$\downarrow$\\  
    
    \hline
    \multirow{2}{*}{a)} Tracking                       & \textcolor{red}{\ding{55}}  &  \textcolor{red}{\ding{55}}  &  \textcolor{red}{\ding{55}} &  \multirow{2}{*}{7.18}    & \multirow{2}{*}{16.76}   & \multirow{2}{*}{0.347}\\
    \multirow{2}{*}{\phantom{a)}} Mapping                        & \textcolor{red}{\ding{55}}  &  \textcolor{red}{\ding{55}}  & \textcolor{red}{\ding{55}}  &       &   \\
    
    \hline 
    \multirow{2}{*}{b)} Tracking                       & \textcolor{red}{\ding{55}}  &  \textcolor{red}{\ding{55}}  &  \textcolor{red}{\ding{55}}     &  \multirow{2}{*}{2.32} & \multirow{2}{*}{19.82}   & \multirow{2}{*}{0.111}\\
    \multirow{2}{*}{\phantom{b)}} Mapping                        & \textcolor{green}\checkmark  & \textcolor{green}\checkmark    &  \textcolor{red}{\ding{55}}  &  & \\
    
    \hline 
    \multirow{2}{*}{c)} Tracking                       & \textcolor{green}\checkmark  & \textcolor{green}\checkmark     &  \textcolor{green}\checkmark & \multirow{2}{*}{6.57} & \multirow{2}{*}{17.25}   & \multirow{2}{*}{0.281}\\
    \multirow{2}{*}{\phantom{c)}} Mapping                        & \textcolor{red}{\ding{55}}   & \textcolor{red}{\ding{55}} & \textcolor{red}{\ding{55}}        &       & \\
    
    \hline 
    \multirow{2}{*}{d)} Tracking                       & \textcolor{green}\checkmark  & \textcolor{green}\checkmark  &  \textcolor{green}\checkmark  & \textbf{\multirow{2}{*}{2.05}}  & \textbf{\multirow{2}{*}{21.23}}  & \textbf{\multirow{2}{*}{0.099}}\\
    \multirow{2}{*}{\phantom{d)}} Mapping                        & \textcolor{green}\checkmark  & \textcolor{green}\checkmark  &  \textcolor{red}{\ding{55}}   &  & \\
    
    \hline 
    \multirow{2}{*}{e)} Tracking                       & \textcolor{green}\checkmark  & \textcolor{green}\checkmark  &  \textcolor{green}\checkmark  & \multirow{2}{*}{2.21}  & \multirow{2}{*}{20.17}  & \multirow{2}{*}{0.115}\\
    \multirow{2}{*}{\phantom{e)}} Mapping                        & \textcolor{green}\checkmark  & \textcolor{green}\checkmark  &  \textcolor{green}\checkmark  &  & \\
    
    \hline
    \end{tabular}
    \label{tab:subtable1}
  \end{subfigure}
  \hfill
  \hspace{20mm} 
  \begin{subfigure}[c]{0.25\linewidth}
    \includegraphics[width=\linewidth]{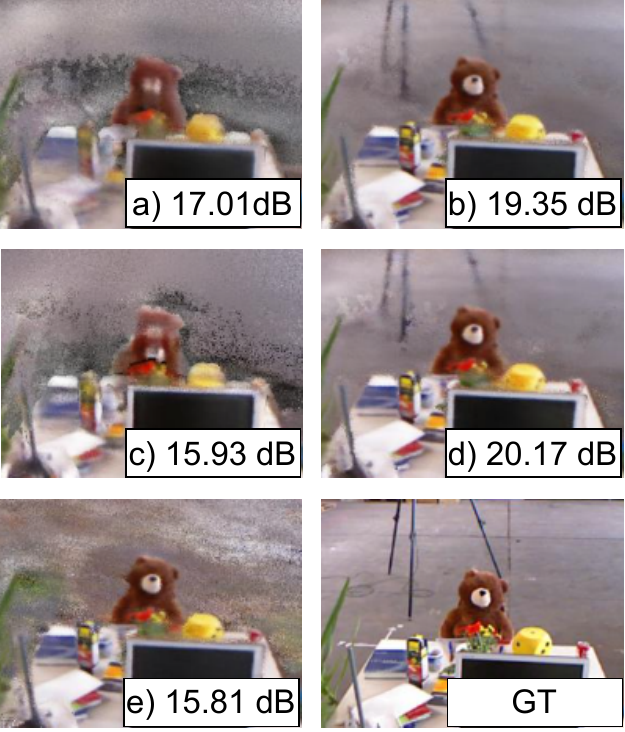} 
    \label{fig:short-3}
  \end{subfigure}
   
  \caption{Ablation on loss term reweighting}
  \label{tab:ablation_reweight}
  \vspace{-3mm}
\end{figure}

\noindent\textbf{Ablation on Reweighting.} 
In \cref{tab:ablation_reweight}, we present a quantitative analysis of the application of model uncertainty to various loss terms on TUM RGB-D \cite{tumrgbd}. Configuration (d) achieves the highest localization accuracy and rendering quality. During tracking, reweighting all terms to focus on only low-uncertainty information improves localization. In mapping, color information can compensate for invalid depth values, so reweighting is not applied to the color term. This strategy enhances reconstruction quality in both geometry (lower depth L1) and appearance (higher PSNR) compared to configuration (e).

\noindent\textbf{Ablation of strategic BA.} \cref{tab:impact_BA} shows localization accuracy and rendering quality under different BA strategies on 6 ScanNet scenes. Experimental results demonstrate that our uncertainty-guided strategic BA method achieves optimal performance by dynamically activating the mapping process and selecting keyframes. \cref{fig:tumrgbd_mesh} ablates the reconstructed mesh without strategic BA.

\vspace{-2mm}

\begin{table}[htbp]
\tiny
\centering
    \begin{tabular}{l | c@{\hspace{7pt}} c@{\hspace{7pt}} c  |   c   |c@{\hspace{7pt}}  c  |c }
    \hline 
    \multirow{2}{*}{Method} & \multicolumn{3}{c|}{Keyframe Selection}  &  Camera   & \multicolumn{2}{c|}{ATE [cm]} &\multirow{2}{*}{PSNR$\uparrow$ }\\
    & Local & Global & LC&  pose &  RMSE$\downarrow$ &  Mean$\downarrow$ \\
    \hline 
    w/o BA            &             &            &            & \ding{55}  & 17.58  & 15.15   &  17.63  \\
    LBA               & \checkmark  &  &            & \checkmark & 8.77   & 7.23    &  20.62  \\
    GBA               &             & \checkmark           &            & \checkmark & 8.35   & 7.17    &  21.52  \\
    LBA + GBA         & \checkmark  & \checkmark &            & \checkmark & 7.23   & 6.56    &  21.59  \\
    LBA + GBA + LLCO  & \checkmark  & \checkmark & \checkmark & \checkmark & \textbf{7.01}  & \textbf{6.15} &  \textbf{21.77}  \\
    \hline
    
    \end{tabular}
\captionsetup{skip=2mm}
\caption{Ablation of strategic BA: LBA selects 20 local keyframes, GBA includes all keyframes, and LLCO focuses on keyframes in loop closure.}
\label{tab:impact_BA}
\vspace{-5mm}
\end{table}

\section{Conclusion}

We present Uni-SLAM, a novel uncertainty-guided dense implicit SLAM approach. In decoupled scene representation, we propose utilizing model-free predictive uncertainty to reweight the loss function at the pixel level to capture effective information, achieving high-frequency geometric reconstruction. By leveraging image-level uncertainty, we strategically perform bundle adjustment to balance local-to-global information. Overall, our method achieves state-of-the-art high-fidelity mapping and accurate tracking in real-time among dense SLAM. 

\noindent We accept a trade-off in efficiency through random sampling in a real-time required SLAM system. However, active sampling based on uncertainty should further improve efficiency and yield finer edge structures. We leave this for future work.

\noindent\textbf{Acknowledgements:} This research has been partially funded by the EU projects CORTEX2 (GA Nr 101070192) and FLUENTLY (GA Nr 101058680).

\clearpage
{\small
\bibliographystyle{ieee_fullname}
\bibliography{citations}
}

\clearpage
\appendix
\setcounter{page}{1}
\maketitlesupplementary
\setcounter{section}{0}
\setcounter{equation}{0}
\renewcommand{\thesection}{\Alph{section}}

\begin{abstract}
In the supplemental material, we provide additional details about the following:
\begin{itemize}
    \item Details on implementation. (Section~\ref{sec:imp_detail})
    \item More analysis and ablation study. (Section~\ref{sec:additional_ablation})
    \item Per-Scene Breakdown of the Results. (Section~\ref{sec:per_scene_resuls})
\end{itemize}
\end{abstract}

\section{Implementation Details}
\label{sec:imp_detail}
\subsection{Hyperparameters}
\noindent \textbf{Default Setting.} For scene representation, we set the hash grid size $L = 16$  for the geometry hash grid and $L=16$ for the appearance hash grid. Default resolutions for both geometry and appearance are $0.02m$. Two tiny 2-layer decoders with $32$ channels are applied to decode the color and the SDF. For the activation functions, ReLU is used in hidden layers, while Sigmoid and Tanh are applied to the output layers for raw color and SDF respectively. We use the Adam optimizer to optimize scene representation and decoder. The learning rate for the geometry hash grid is $5e^{-2}$, the learning rate for the appearance hash grid is also $5e^{-2}$, and the learning rate for both MLP decoders is $5e{-3}$. We sample $N_{str} = 32$ stratified points and $N_{imp} = 10$ points within the truncated distance $\tau_{tr} = 6cm$. Our pixel-level uncertainty threshold is $\beta_{unc_m} = 1e{-2}$, image-level uncertainty threshold is $\beta_{unc} = 1e{-3}$ and the co-visibility threshold is ${OC}_{cov} = 0.95$. We always optimize the camera pose during tracking and mapping if BA is enabled. The learning rate for camera pose rotation and translation is $1e{-3}$. The weights of the loss function are $\lambda_{rgb}=5$, $\lambda_{dep} = 0.1$, $\lambda_{sdf_c}=200$, $\lambda_{sdf_t}=10$ and $\lambda_{sdf_{fs}}=5$ for mapping, while $\lambda_{rgb}=5$, $\lambda_{dep} = 1$, $\lambda_{sdf_c}=200$, $\lambda_{sdf_t}=50$ and $\lambda_{sdf_{fs}}=10$ are set for tracking. For the tracking part, we perform the tracking process for every frame, select $M_t = 2000$ sampling points, and perform 8 iterations. For the mapping part, we select $M_m = 4000$ sampling points, perform 13 iterations every 4 frames and use a window of $W = 20$ keyframes for local bundle adjustment. At the start of training,  we use 200 iterations for the first frame mapping. The reconstructed mesh is extracted by marching cubes algorithm\cite{marchingcube}. To ensure a fair comparison, we do the same mesh culling strategy for all benchmark baselines following Neural-RGBD\cite{neuralrgbd}. In order to present the reconstructed quality considering both tracking and mapping, the predicted camera poses are used for culling paths instead of ground truth poses.

\noindent \textbf{Replica Dataset \cite{replica}} We set $L=19$ for the appearance hash grid. Replica dataset it contains eight synthetic scenes including 3D ground truth mesh. So based on its 3D ground truth mesh we can also evaluate our metrics on 3D evaluation, such as \emph{Depth L1 [cm], Accuracy [cm], Reconstruction completion [cm]},  and \emph{Completion ratio [$<$ 1cm ]\%}. Those meshes are culled following \cite{neuralrgbd} before evaluation. 

\noindent \textbf{ScanNet Dataset \cite{scannet}} We perform the mapping process every 5 frames, increasing the number of iterations to 20, $N_{str} = 48$. For tracking, iterations are increased to 20. Because of invalid depth at the edge of the image of ScanNet, 75 pixels are culled at the edge of the image for tracking during data pre-processing. The learning rate of translation is set to $5e^{-4}$, and the learning rate of rotation is $3e^{-3}$. 

\noindent \textbf{TUM RGB-D Dataset \cite{tumrgbd}}
The image-level uncertainty threshold is increased to $\beta_{unc} = 2e{-3}$. We perform a mapping process every 4 frames here and select $M=4000$ sampling points for tracking and mapping. 20 pixels are culled at the edge of the image for tracking. The iteration of tracking is set to 20, while the iteration of mapping is also set to 20, $N_{str} = 48$. The learning rate of two hash grids is set to $2e^{-2}$. The learning rate of translation is set to $1e^{-2}$, and the learning rate of rotation is $5e^{-3}$. 
\subsection{Proof of Termination Probability}
Our goal is to prove the accumulated termination probability along a current sampling ray $r$ as:
$$
p(r)=\sum_{n=1}^N w_n=1
$$
where $N$ is the number of sampling points along the ray $r$, the weight $w_n$ is defined as:
$$
w_n=T_n \cdot\left(1-\exp \left(-\sigma\left(p_n\right)\right)\right)
$$
where $p_n$ is one sampling point along this ray, $T_n$ is the transmittance of all previous sample points.
$$
T_n=\exp \left(-\sum_{k=1}^{n-1} \sigma\left(p_k\right)\right)
$$

First, we expand the weight $w_n$:
$$
\sum_{n=1}^N w_n=\sum_{n=1}^N\left(\exp \left(-\sum_{k=1}^{n-1} \sigma\left(p_k\right)\right) \cdot\left(1-\exp \left(-\sigma\left(p_n\right)\right)\right)\right)
$$

Second, introduce a recursive relationship for transmittance. We know that the relationship between $T_n$ and $T_{n+1}$ is:
$$
T_{n+1}=T_n \cdot \exp \left(-\sigma\left(p_n\right)\right)
$$

So we can expand term by term and see the pattern:
$$
\begin{aligned}
& T_1=1 \\
& T_2=\exp \left(-\sigma\left(p_1\right)\right) \\
& T_3=\exp \left(-\sigma\left(p_1\right)\right) \cdot \exp \left(-\sigma\left(p_2\right)\right)=\exp \left(-\sigma\left(p_1\right)-\sigma\left(p_2\right)\right)
\end{aligned}
$$

Thus, for any $n$ :
$$
T_n=\exp \left(-\sum_{k=1}^{n-1} \sigma\left(p_k\right)\right)
$$

According to Equation:
$$
\sum_{n=1}^N w_n=\sum_{n=1}^N\left(\exp \left(-\sum_{k=1}^{n-1} \sigma\left(p_k\right)\right) \cdot\left(1-\exp \left(-\sigma\left(p_n\right)\right)\right)\right)
$$

Look at it item by item:
$$
\begin{aligned}
w_1 &= T_1 \cdot \left(1 - \exp\left(-\sigma(p_1)\right)\right) \\
    &= 1 \cdot \left(1 - \exp\left(-\sigma(p_1)\right)\right) \\
    &= 1 - \exp\left(-\sigma(p_1)\right) \\
w_2 &= T_2 \cdot \left(1 - \exp\left(-\sigma(p_2)\right)\right) \\
    &= \exp\left(-\sigma(p_1)\right) \cdot \left(1 - \exp\left(-\sigma(p_2)\right)\right) \\
    &= \exp\left(-\sigma(p_1)\right) - \exp\left(-\sigma(p_1) - \sigma(p_2)\right) \\
w_3 &= T_3 \cdot \left(1 - \exp\left(-\sigma(p_3)\right)\right) \\
    &= \exp\left(-\sigma(p_1) - \sigma(p_2)\right) \cdot \left(1 - \exp\left(-\sigma(p_3)\right)\right) \\
    &= \exp\left(-\sigma(p_1) - \sigma(p_2)\right) - \exp\left(-\sigma(p_1) - \sigma(p_2) - \sigma(p_3)\right)
\end{aligned}
$$

Continuing in this way, we can discover the structure of each item:

\begin{align*}
\sum_{n=1}^N w_n= & \left(1-\exp \left(-\sigma\left(p_1\right)\right)\right) \\
& +\left(\exp \left(-\sigma\left(p_1\right)\right)-\exp \left(-\sigma\left(p_1\right)-\sigma\left(p_2\right)\right)\right) \\
& +\left(\exp \left(-\sigma\left(p_1\right)-\sigma\left(p_2\right)\right) \right.\\
& \left.-\exp \left(-\sigma\left(p_1\right)-\sigma\left(p_2\right)-\sigma\left(p_3\right)\right)\right) \\
& +\cdots \\
& +\left(\exp \left(-\sum_{k=1}^{N-1} \sigma\left(p_k\right)\right)-\exp \left(-\sum_{k=1}^N \sigma\left(p_k\right)\right)\right)
\end{align*}

All the intermediate terms cancel each other out, leaving only the first and last terms:
$$
\sum_{n=1}^N w_n=1-\exp \left(-\sum_{k=1}^N \sigma\left(p_k\right)\right)
$$

As $N$ tends to infinity, assuming all densities are cumulative in the observed regions, the exponential part of the last term tends to negative infinity, then:
$$
\exp \left(-\sum_{k=1}^N \sigma\left(p_k\right)\right) \approx 0
$$

So,
$$
\sum_{n=1}^N w_n=1-0=1
$$

By the above steps, we have proved that the cumulative sum of all weights $w_n$ on a ray for observed area is equal to 1. However, in unobserved regions where the density values $\sigma(p_k)$ are very small or zero, the exponential term will tend to 1, so

$$\sum_{n=1}^N w_n=1-1=0 $$
Therefore, the termination probability is proven to lie within the range $(0,1)$.

\subsection{Co-visibility Check}
Loop detection is implemented based on sample point remapping. We sample $M=50$ pixels for every keyframe in the keyframes database, sample $N=8$ sample points along each ray, given the camera's internal and external parameters, and map these points back to the current frame. If the overlap coefficient is greater than $0.95$, we consider that a loop closure has occurred. In order to avoid too short a time interval and too short a range of motion for loop closure detection, we set a minimum threshold of 100 frames between the two points where a loop closure occurs.

\section{More Analysis and Ablation Study}
\label{sec:additional_ablation}

\subsection{Hash Grid Size Analysis}
To investigate the distinct requirements of geometry and appearance for spatial representation, we conduct our experiments on the synthetic Replica dataset. We evaluate different hash grid size combinations to investigate the sensitivity of appearance and geometry to hash grid size in \cref{tab:hash_grid_size_psnr_comp}, while \cref{tab:hash_grid_size_modelsize_fps} compares the impact of hash grid size on model size and speed in frame per second(FPS). We compare the results with BSLAM\cite{birn-slam},Co-SLAM\cite{coslam} and ESLAM\cite{eslam} at index 3, index 5, index 7 respectively. In these plots, the numbers in parentheses $(h_g,h_a)$ report the geometry hash grid size and appearance hash grid size respectively. Experiments show that the reconstruction and rendering quality can be further improved by increasing the hash grid size. However, for equal model sizes, allocating more memory to appearance yields more benefits on rendering quality and completeness (compare the combination of index 4 $(h_g=16,h_a=19)$ and index 9 $(h_g=19,h_a=16)$). We interpret this phenomenon by considering that \textit{color information is a higher-frequency signal compared to geometric information}. The implication here is that when computational resources are limited, we should allocate more resources to the appearance signal. In terms of the relation between hash grid size and FPS, it is worth noting that when increasing the hash grid size combination from $(h_g=16,h_a=19)$ to $(h_g=22,h_a=22)$, the speed in FPS only decreases from 8.3 fps to 6.6 fps. 

\begin{table}[ht]
\centering
\vspace{0mm}
\captionsetup{skip=10pt}
\begin{tikzpicture}
    \begin{groupplot}[
        group style={
            group size=1 by 2,
            vertical sep=1.5cm   
        },
        width=7cm,  
        height=4.4cm,
        title style={font=\scriptsize},
        xlabel style={font=\scriptsize},
        ylabel style={font=\scriptsize, yshift=-5mm},
        legend style= {font=\fontsize{5}{6}\selectfont, inner sep=0.1pt, outer sep=0.1pt},
        tick label style={font=\tiny},
        grid style=dashed,
    ]

    \nextgroupplot[
        title={(a) Impact on PSNR.},
        xlabel={Index},
        ylabel={PSNR(dB)},
        xmin=0, xmax=12,
        ymin=29, ymax=32.5,
        legend pos=south east,
        xlabel style={yshift=0mm}, 
        ylabel style={yshift=3mm}, 
        xtick={1,2,3,4,5,6,7,8,9,10,11}
    ]
    \addplot[color=green, mark=triangle*] coordinates {(1,30.33)(2,30.59)(3,31.02)(4,31.62)(5,31.25)(6,31.38)(7,31.77)(8,31.07)(9,31.82)(10,31.75)(11,31.95)};
    \addlegendentry{Ours}
    \addplot[color=blue, mark=*, only marks] coordinates {(7,30.19)};
    \addlegendentry{ESLAM}
    \addplot[color=red, mark=*, only marks] coordinates {(5,30.27)};
    \addlegendentry{Co-SLAM}
    \addplot[color=black, mark=*, only marks] coordinates {(3,29.55)};
    \addlegendentry{BSLAM}
    \node at (axis cs:1,30.33) [anchor=south] {\texttt{\tiny{(14,14)}}};
    \node at (axis cs:2,30.59) [anchor=south] {\texttt{\tiny{(15,15)}}};
    \node at (axis cs:3,31.02) [anchor=south] {\texttt{\tiny{(16,16)}}};
    \node at (axis cs:4,31.62) [anchor=south] {\texttt{\tiny{(16,19)}}};
    \node at (axis cs:5,31.25) [anchor=south] {\texttt{\tiny{(17,17)}}};
    \node at (axis cs:6,31.38) [anchor=south] {\texttt{\tiny{(18,18)}}};
    \node at (axis cs:7,31.77) [anchor=south] {\texttt{\tiny{(19,19)}}};
    \node at (axis cs:8,31.07) [anchor=south] {\texttt{\tiny{(19,16)}}};
    \node at (axis cs:9,31.82) [anchor=south] {\texttt{\tiny{(20,20)}}};
    \node at (axis cs:10,31.75) [anchor=north] {\texttt{\tiny{(21,21)}}};
    \node at (axis cs:11,31.95) [anchor=north] {\texttt{\tiny{(22,22)}}};

    \nextgroupplot[
        title={(b) Impact on Completion Rate.},
        xlabel={Index},
        ylabel={Completion Ratio (\%)},
        xmin=0, xmax=12,
        ymin=55, ymax=70,
        legend pos=south east,
        xlabel style={yshift=0mm}, 
        ylabel style={yshift=3mm}, 
        xtick={1,2,3,4,5,6,7,8,9,10,11},
    ]
    \addplot[color=green, mark=triangle*] coordinates {(1,58.72)(2,61.33)(3,63.57)(4,67.09)(5,64.89)(6,66.23)(7,67.34)(8,65.23)(9,67.56)(10,67.63)(11,67.61)};
    \addlegendentry{Ours}
    \addplot[color=blue, mark=*, only marks] coordinates {(7, 62.25)};
    \addlegendentry{ESLAM}
    \addplot[color=red, mark=*, only marks] coordinates {(5,55.94)};
    \addlegendentry{Co-SLAM}
    \addplot[color=black, mark=*, only marks] coordinates {(3,57.18)};
    \addlegendentry{BSLAM}
    \node at (axis cs:1,58.72) [anchor=south] {\texttt{\tiny{(14,14)}}};
    \node at (axis cs:2,61.33) [anchor=south] {\texttt{\tiny{(15,15)}}};
    \node at (axis cs:3,63.57) [anchor=south] {\texttt{\tiny{(16,16)}}};
    \node at (axis cs:4,67.09) [anchor=south] {\texttt{\tiny{(16,19)}}};
    \node at (axis cs:5,64.89) [anchor=south] {\texttt{\tiny{(17,17)}}};
    \node at (axis cs:6,66.23) [anchor=south] {\texttt{\tiny{(18,18)}}};
    \node at (axis cs:7,67.34) [anchor=south] {\texttt{\tiny{(19,19)}}};
    \node at (axis cs:8,65.23) [anchor=south] {\texttt{\tiny{(19,16)}}};
    \node at (axis cs:9,67.56) [anchor=south] {\texttt{\tiny{(20,20)}}};
    \node at (axis cs:10,67.63) [anchor=north, yshift=0mm] {\texttt{\tiny{(21,21)}}};
    \node at (axis cs:11,67.61) [anchor=south] {\texttt{\tiny{(22,22)}}};

    \end{groupplot}
\end{tikzpicture}
\caption{Impact of \texttt{(SDF hash grid size,  Appearance hash grid size)} on PSNR [dB] and Completion Rate [$<1cm \%$] on the Replica dataset.} 
\label{tab:hash_grid_size_psnr_comp}
\end{table}

\begin{table}[ht]
\centering
\begin{tikzpicture}
    \begin{groupplot}[
        group style={
            group size=1 by 2,
            vertical sep=1.5cm   
        },
        width=7cm,  
        height=4.4cm,
        title style={font=\scriptsize},
        xlabel style={font=\scriptsize},
        ylabel style={font=\scriptsize, yshift=-5mm},
        legend style= {font=\fontsize{5}{6}\selectfont, inner sep=0.1pt, outer sep=0.1pt},
        tick label style={font=\tiny},
        grid style=dashed,
    ]

    \nextgroupplot[
        title={(a) Impact on Model Size},
        xlabel={Index},
        ylabel={Model Size (MB)},
        xmin=0, xmax=10,
        ymin=0, ymax=44,
        legend pos=north west,
        xlabel style={yshift=0mm}, 
        ylabel style={yshift=3mm}, 
        xtick={1,2,3,4,5,6,7,8,9}
    ]
    \addplot[color=green, mark=triangle*] coordinates {(1,0.97)(2,1.84)(3,3.47)(4,12.69)(5,6.48)(6,12.06)(7,12.69)(8,22.35)(9,41.15)};
    \node at (axis cs:1,0.97) [anchor=south] {\texttt{\tiny{(14,14)}}};
    \node at (axis cs:2,1.84) [anchor=south] {\texttt{\tiny{(15,15)}}};
    \node at (axis cs:3,3.47) [anchor=south] {\texttt{\tiny{(16,16)}}};
    \node at (axis cs:4,12.69) [anchor=south] {\texttt{\tiny{(16,19)}}};
    \node at (axis cs:5,6.48) [anchor=north] {\texttt{\tiny{(17,17)}}};
    \node at (axis cs:6,12.06) [anchor=north] {\texttt{\tiny{(18,18)}}};
    \node at (axis cs:7,12.69) [anchor=south] {\texttt{\tiny{(19,16)}}};
    \node at (axis cs:8,22.35) [anchor=south] {\texttt{\tiny{(19,19)}}};
    \node at (axis cs:9,41.15) [anchor=east, yshift=-2mm] {\texttt{\tiny{(20,20)}}};

    \addlegendentry{Ours}
    \addplot[color=blue, mark=*, only marks] coordinates {(7, 6.78)};
    \addlegendentry{ESLAM}
    \addplot[color=red, mark=*, only marks] coordinates {(5,1.72)};
    \addlegendentry{Co-SLAM}
    \addplot[color=black, mark=*, only marks] coordinates {(3,17.38)};
    \addlegendentry{BSLAM}
    
    \nextgroupplot[
        title={(b) Impact on FPS.},
        xlabel={Index},
        ylabel={FPS},
        xmin=0, xmax=10,
        ymin=0, ymax=10,
        legend pos=south east,
        xlabel style={yshift=0mm}, 
        ylabel style={yshift=3mm}, 
        xtick={1,2,3,4,5,6,7,8,9},
    ]
    \addplot[color=green, mark=triangle*] coordinates {(1,8.6)(2,8.6)(3,8.6)(4,8.5)(5,8.4)(6,8.3)(7,8.3)(8,7.57)(9,6.6)};
    \node at (axis cs:1,8.6) [anchor=south] {\texttt{\tiny{(14,14)}}};
    \node at (axis cs:2,8.6) [anchor=north] {\texttt{\tiny{(15,15)}}};
    \node at (axis cs:3,8.6) [anchor=south] {\texttt{\tiny{(16,16)}}};
    \node at (axis cs:4,8.5) [anchor=north] {\texttt{\tiny{(16,19)}}};
    \node at (axis cs:5,8.4) [anchor=south] {\texttt{\tiny{(17,17)}}};
    \node at (axis cs:6,8.3) [anchor=north] {\texttt{\tiny{(18,18)}}};
    \node at (axis cs:7,8.3) [anchor=south] {\texttt{\tiny{(19,16)}}};
    \node at (axis cs:8,7.57) [anchor=south] {\texttt{\tiny{(19,19)}}};
    \node at (axis cs:9,6.6) [anchor=east] {\texttt{\tiny{(20,20)}}};

    \addlegendentry{Ours}
    \addplot[color=blue, mark=*, only marks] coordinates {(7, 5.55)};
    \addlegendentry{ESLAM}
    \addplot[color=red, mark=*, only marks] coordinates {(5,9.07)};
    \addlegendentry{Co-SLAM}
    \addplot[color=black, mark=*, only marks] coordinates {(3,1.66)};
    \addlegendentry{BSLAM}

    \end{groupplot}
\end{tikzpicture}
\caption{Impact of \texttt{(SDF hash grid size, Appearance hash grid size)} on Model Size and FPS on the Replica dataset.} 
\label{tab:hash_grid_size_modelsize_fps}
\end{table}

\subsection{Strategic BA Analysis}
\begin{figure}[htbp]
  \centering
  \includegraphics[height=7cm]{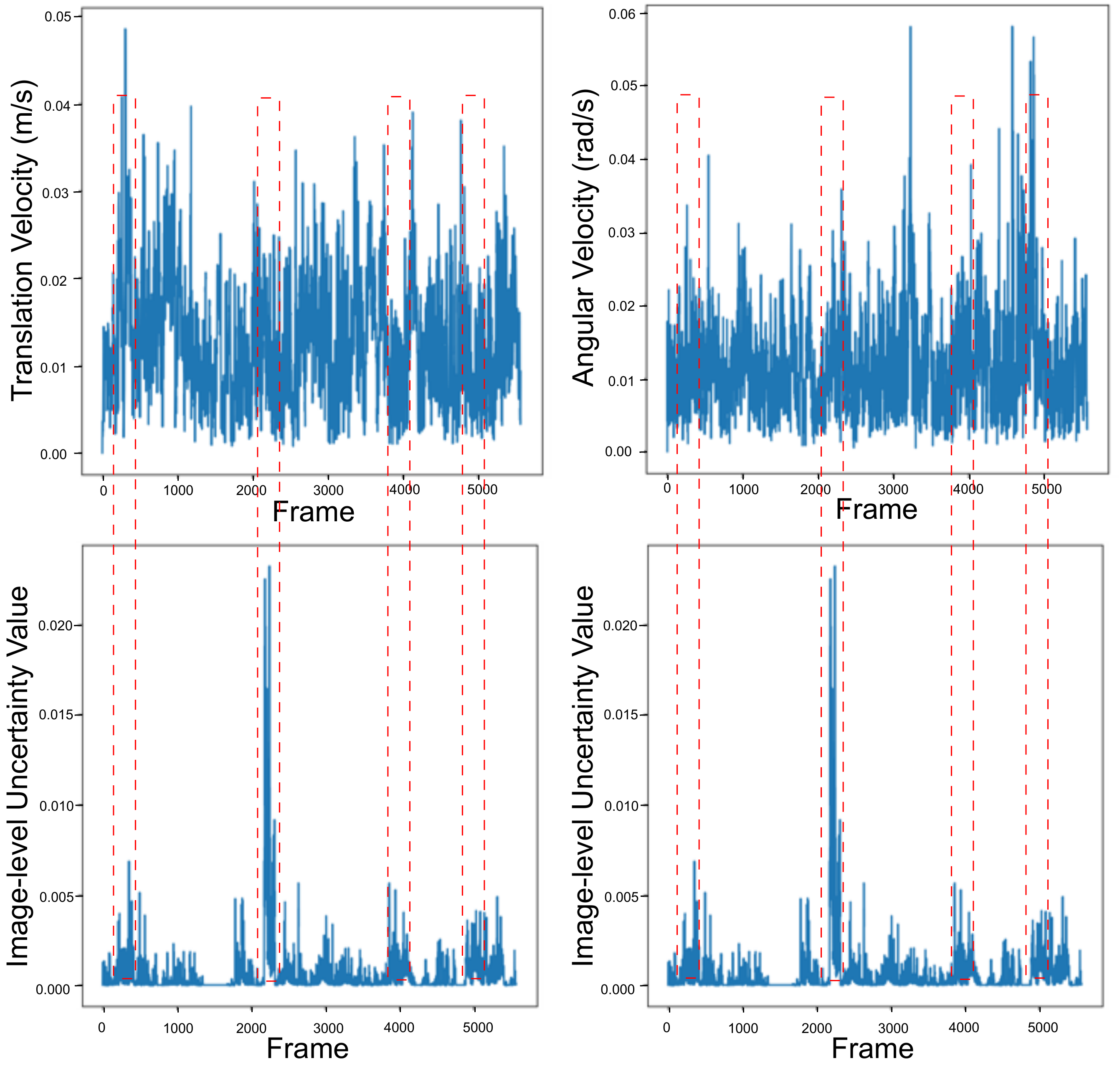}
  \caption{\textbf{Impact of Translational and Angular Velocities on Uncertainty.} We can observe the correlation between uncertainty and both translational velocity and angular velocity. Higher velocities or accelerations tend to result in higher uncertainty.}
  \label{fig:velocity_angle_speed_uncertainty}
\end{figure}
\noindent \textbf{Uncertainty vs. Velocity.} To analyze the relationship between velocity and uncertainty, we conducted experiments on \texttt{scene0000} from ScanNet \cite{scannet}. The camera's motion state is described in terms of translation and rotation $\{T_i | R_i\}$. In \cref{fig:velocity_angle_speed_uncertainty}, we visualize the translational velocity and angular velocity, with the corresponding image-level uncertainty displayed below each. The results show that higher velocities or accelerations can easily cause the camera to move into unseen areas before, leading to increased uncertainty. This figure exposes the relationship between our definition of uncertainty and the state of camera motion, justifying our definition of uncertainty.

\begin{figure}[htbp]
  \centering
  \includegraphics[height=5cm]{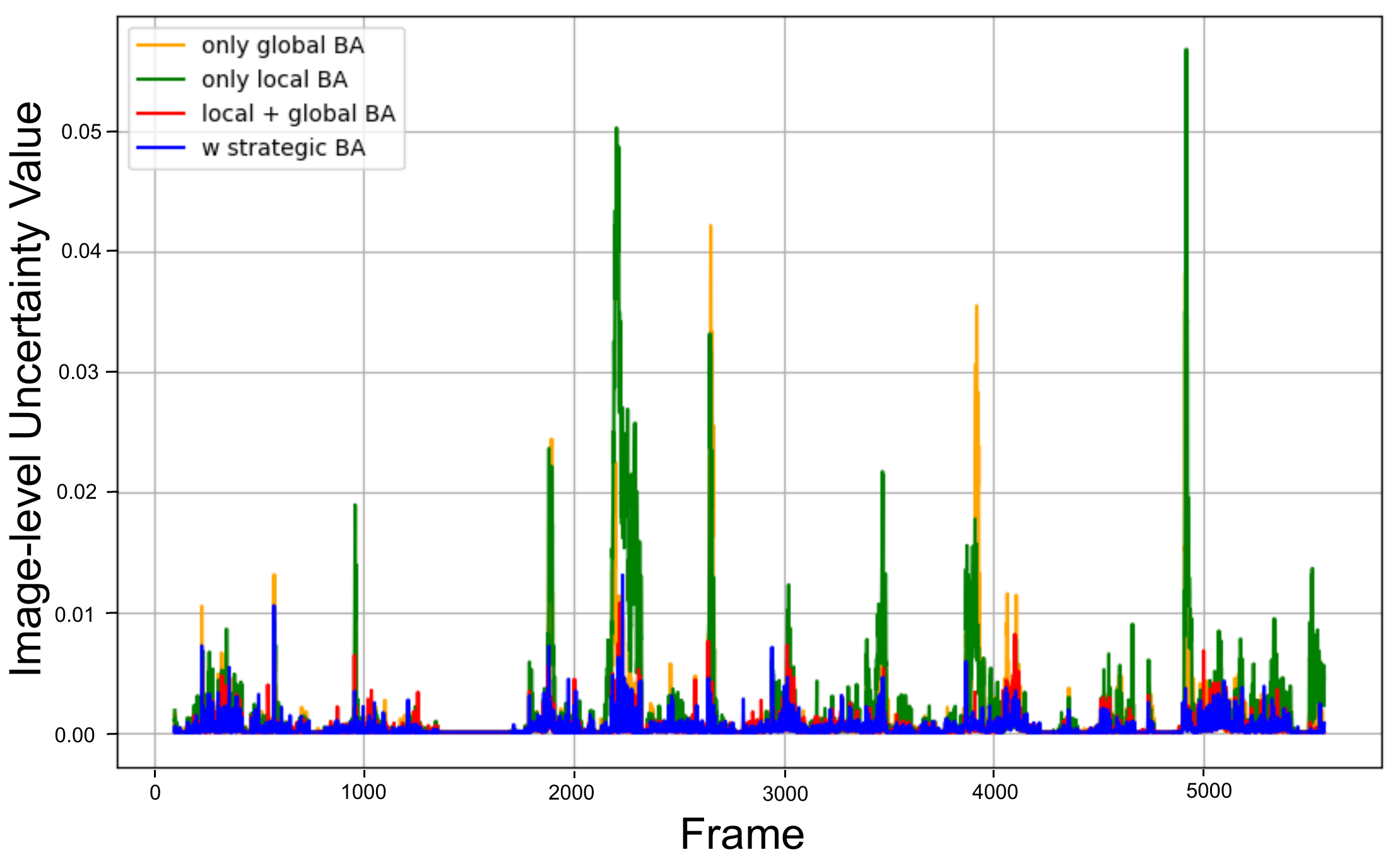}
  \caption{\textbf{Impact of different keyframe selection on Uncertainty.} Here we compare the changing image-level uncertainty per frame with different keyframe selection strategies. The results indicated by the \textcolor{blue}{blue} line show that image-level uncertainty is significantly reduced, achieving optimal outcomes with our proposed strategic BA (local BA + global BA + LLCO).}
  \label{fig:strategic_BA_comparison_chart}
\end{figure}

\begin{figure*}[htbp]
  \centering
  \includegraphics[height=6cm]{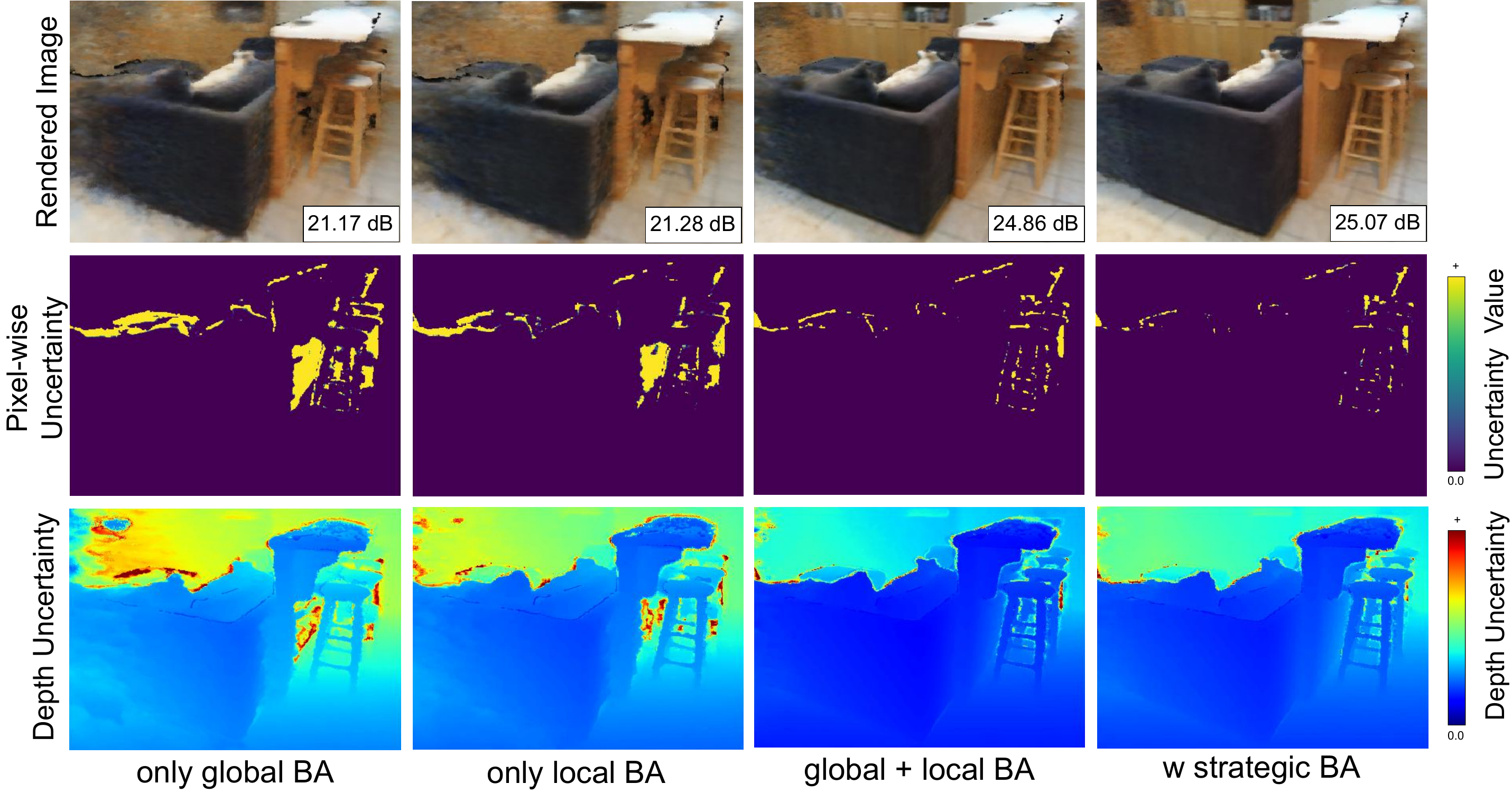}
  \caption{\textbf{Impact of Strategic BA on Rendering and Uncertainty Visualization.} Our proposed strategic BA integrates global BA, local BA, and LLCO. This approach achieves the highest rendered image quality, as indicated by the PSNR (dB) metric. The second row presents visualized pixel-level uncertainty, while depth uncertainty illustrates geometric reconstruction in the third row. The depth uncertainty, defined in \cref{eq:depth_uncertainty}, shows a continuous variation in visualized uncertainty, providing a clearer demonstration of the superiority of our approach.}
  \label{fig:strategic_BA_comparison_fig}
\end{figure*}

\noindent \textbf{Impact of Strategic BA on Uncertainty.} We investigated the impact of using strategic Bundle Adjustment (BA) on image-level uncertainty on \texttt{scene0000} from ScanNet \cite{scannet}. 
As shown in \cref{fig:strategic_BA_comparison_chart}, using only constant global BA results in high uncertainty, as indicated by the \textcolor{orange}{orange} line. Similarly, the \textcolor{green}{green} line represents high uncertainty with only local BA. The \textcolor{red}{red} line shows suboptimal results when using global BA and local BA without local loop closure optimization (LLCO).
However, with our full strategic BA the uncertainty could be reduced significantly on average as shown in \textcolor{blue}{blue} line. This implies more accurate localization and improved rendering. Further reduction in uncertainty demonstrates the effectiveness of our LLCO approach. In \cref{fig:strategic_BA_comparison_fig}, we present the visual results. We visualize rendered image, depth uncertainty, and pixel-level uncertainty in three rows respectively. It is evident that under strategic BA, the quality of rendered images is noticeably enhanced, and the corresponding depth uncertainty is also lower, indicating higher geometric quality. The depth uncertainty is calculated as follows:
\begin{equation}
\hat{d}_{unc}=\sqrt{\sum_{i=1}^N w_i\left(\hat{\boldsymbol{d}}-d_i\right)^2}
\label{eq:depth_uncertainty}
\end{equation}
where $w_i$ is the weight corresponding to Equation(2) in main paper, $\hat{\boldsymbol{d}}$ is predicted depth, and $d_i$ represents the distance from the camera center to the current sample point $\mathbf{x}_i$ along this ray. Pixel-level uncertainty in the third column is also lower with this strategy.

\noindent \textbf{Plug-in Capability.} The effectiveness of our strategy has also been validated on BSLAM\cite{birn-slam}. Based on image-level uncertainty and co-visibility check, we dynamically activate an additional mapping process beyond the global BA. The results in \cref{tab:strategic_BA_birn_slam} show improvements in all metrics, benefiting from our uncertainty-aware strategy. This demonstrates the plug-in capability of our approach.

\begin{table}[ht]
\scriptsize
\centering
\caption{Analysis of the impact of our strategic BA on BSLAM \cite{birn-slam} (Sec. 3.4 in the main paper). The experiment is conducted on Replica \cite{replica}, and the metrics are ATE RMSE (cm), reconstruction accuracy (cm), reconstruction completion (cm), completion ratio and PSNR. BSLAM \cite{birn-slam} can also benefit from our strategy.}

\begin{tabular}{l | cccc}
\toprule
Method  & ATE & Acc.  & Comp. Ratio  & PSNR  \\
 & (cm)$\downarrow$ &  $ (cm) \downarrow$ & $[<1cm \%] \uparrow$  & (dB) $\uparrow$  \\
 
\midrule
BSLAM \cite{birn-slam}    & 1.19  & 1.12  & 57.18 & 29.55\\
BSLAM w/ Our strategic BA & 1.07 & 1.01  & 58.36 & 29.83\\
\bottomrule
\end{tabular}
\label{tab:strategic_BA_birn_slam}
\end{table}

\subsection{Ablation on Model Design}
\begin{figure}[htbp]
  \centering
  \includegraphics[height=4.3cm]{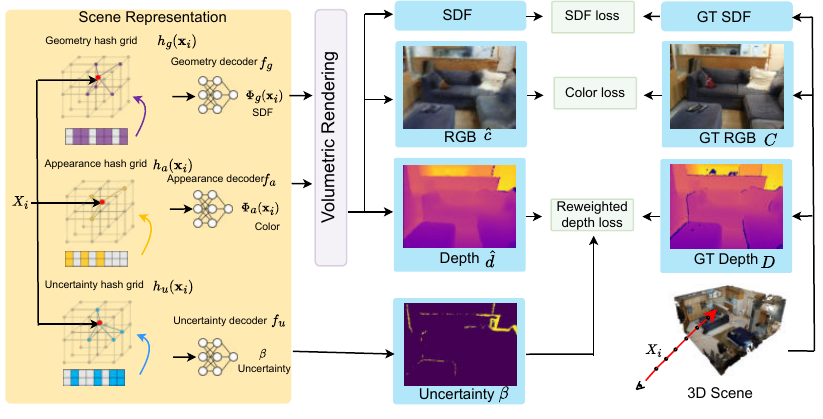}
  \caption{\textbf{Ablation on Gaussian Assumption Uncertainty Model.} We use three grids to represent geometry, appearance, and learnable uncertainty respectively.}
  \label{fig:learnable_unc_pipeline}
\end{figure}

In order to justify our choice of a model-free uncertainty model, we conduct also experiments with a learnable uncertainty model. As shown in \cref{fig:learnable_unc_pipeline}, in addition to using two sparse grids to represent geometry and appearance separately, we use a third grid to model depth uncertainty based on the Gaussian assumption inspired by \cite{naruto}. For depth uncertainty, a model posterior assumption is made from the Bayesian perspective, similar to Bayes' Rays \cite{bayesrays}. Our experiments show that this idea not only brings undesirable increased model complexity, making the model much slower, but also leads to poorer results in terms of reconstruction quality (corresponding to main paper Sec. 4.3). 

The following paragraph explains how we design learnable uncertainty to reweight the depth term loss function.

\noindent\textbf{Gaussian Assumption Uncertainty:} 
Assume that the residuals (errors) between the estimated depth $\hat{d}$ and the true depth $D$ follow a Gaussian distribution with variance $\sigma^2$:

\begin{equation}
    \hat{d} \sim \mathcal{N}(D, \sigma^2)
\end{equation}

The probability density function (PDF) of a normal distribution is given by:
\begin{equation}
p(\hat{d} | D, \sigma^2) = \frac{1}{\sqrt{2\pi\sigma^2}} \exp \left( -\frac{(\hat{d} - D)^2}{2\sigma^2} \right)
\end{equation}

\noindent To maximize the likelihood, we equivalently minimize the negative log-likelihood. The negative log-likelihood for a single observed ray is given by:

\begin{equation}
-\log p(\hat{d} | D, \sigma^2) = \frac{(\hat{d} - D)^2}{2\sigma^2} + \frac{1}{2} \log (2\pi\sigma^2)
\end{equation}

For simplicity, we often drop the constant term $\frac{1}{2} \log (2\pi)$ since it does not affect the optimization. Here we let $\beta = \sigma^2$. In practice, we work with an estimate of the variance $\beta$ through a third grid parallel with the geometry and appearance grid. So, the term we need to minimize is:

\begin{equation}
\mathcal{L}_{\text{single}} = \frac{(\hat{d} - D)^2}{2\sigma^2} + \frac{1}{2} \log \sigma^2 = \frac{(\hat{d} - D)^2}{2\beta} + \frac{1}{2} \log \beta
\end{equation}

If we have a set of depth measurements $R_d$, we sum the negative log-likelihoods for all rays $r$ in the set $R_d$. Additionally, we normalize by the number of elements $|R_d|$ to get the average loss:

This matches the given loss function:

\begin{equation}
\mathcal{L}_d = \frac{1}{|R_d|} \sum_{r \in R_d} \left( \frac{1}{2\beta} (\hat{d}_r - D_r)^2 + \frac{1}{2} \log \beta \right)
\end{equation}

The first term $\frac{(\hat{d} - D)^2}{2\beta}$ penalizes large errors more if the predicted uncertainty $\beta$ is small. The second term $\frac{1}{2} \log \beta$ prevents the model from predicting an arbitrarily large uncertainty to minimize the first term. By balancing these two terms, the loss function encourages the model to provide both accurate depth estimates and reasonable uncertainty estimates.

\begin{figure}[htbp]
  \centering

    \scriptsize
    \begin{tabular}{l | c@{\hspace{3pt}} | c@{\hspace{3pt}}  c@{\hspace{3pt}}   |c@{\hspace{3pt}}  c@{\hspace{5pt}} c@{\hspace{3pt}}}
    \hline 
    \multirow{2}{*}{Dataset} & \multirow{2}{*}{Method}   & \multicolumn{2}{c|}{Tracking/Rendering} & \multirow{2}{*}{FPS $\uparrow$}  & \multirow{2}{*}{params $\downarrow$}\\
                            &                            & \tiny RMSE [cm] $\downarrow$ & \tiny PSNR [dB] $\uparrow$  \\  
    
    \hline
    \multirow{2}{*}{Replica\cite{replica}} & Gaussian   & 1.175  & 27.27      &  7.06     & 14.65M \\
                                           & Ours       & \textbf{0.45}  & \textbf{31.62}     & \textbf{8.37}    & \textbf{12.69M}\\

    \hline
    \multirow{2}{*}{ScanNet\cite{scannet}} & Gaussian   & 11.93 & 18.06    &   3.57    &  5.13M \\
                                           & Ours       & \textbf{7.01}   & \textbf{21.77}     & \textbf{4.88}    &  \textbf{3.39M} \\

    \hline
    \multirow{2}{*}{TUM RGB-D\cite{tumrgbd}} & Gaussian   & 2.16  & 19.30   &   1.73     & 5.38M        \\
                                             & Ours       & \textbf{2.05}   & \textbf{21.23}  & \textbf{2.72}     & \textbf{3.58M}  \\

    \hline
    \end{tabular}

  \caption{\textbf{Gaussian Assumption Model vs. Ours.} }
  \label{tab:ablation_model_design}
  \vspace{-3mm}
\end{figure}

\begin{figure}[htbp]
  \centering
  \includegraphics[height=15cm]{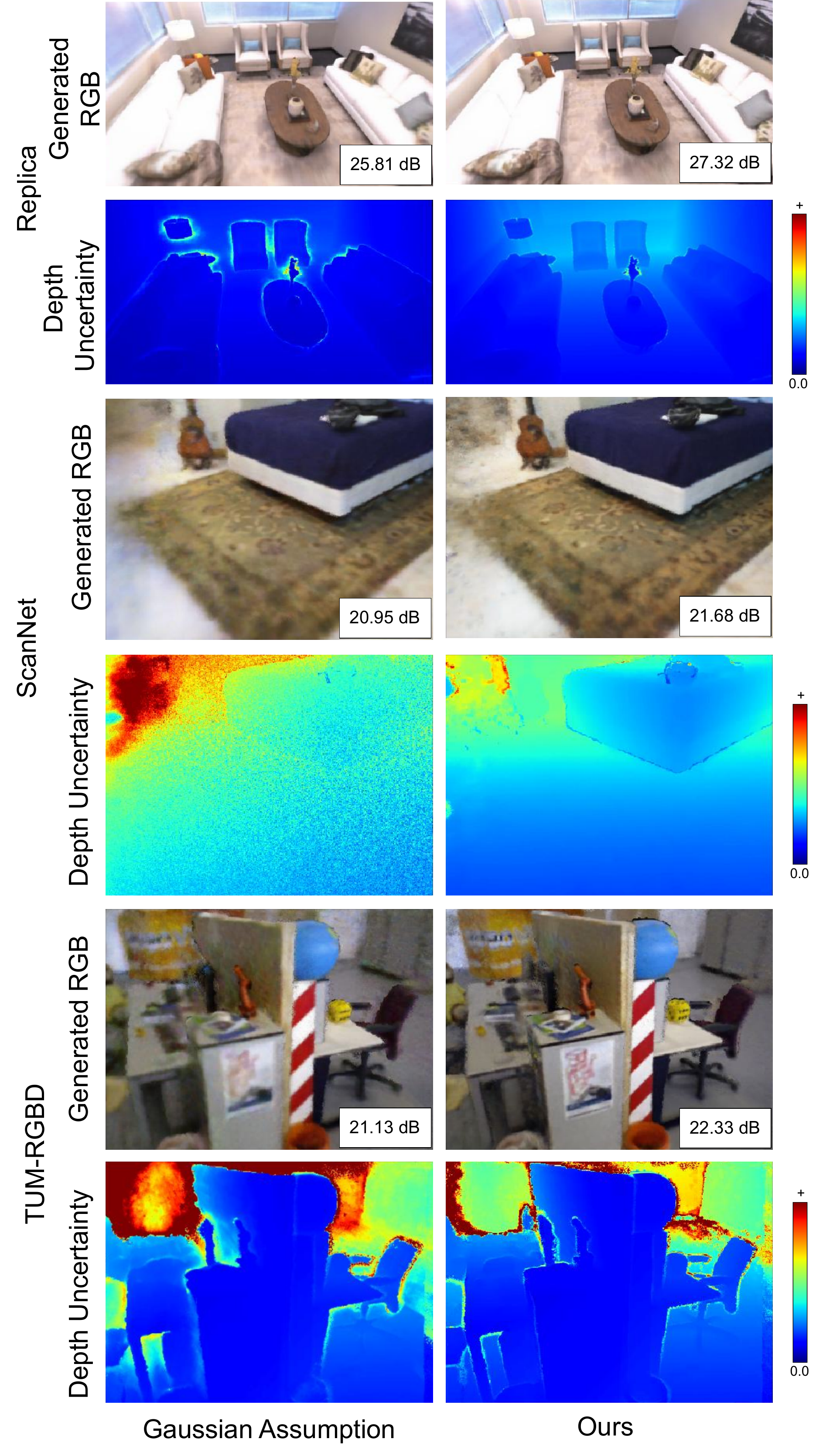}
  \caption{\textbf{Gaussian Assumption Model vs. Ours.} Our model demonstrates superior rendering quality, as evaluated by PSNR (dB) $\uparrow$. Depth uncertainty, calculated using \cref{eq:depth_uncertainty}, is visualized for comparison. Our method visibly reduces depth uncertainty, as clearly shown in the visualizations.}
  \label{fig:gaussian_vs_us}
\end{figure}

Our system demonstrates superior tracking accuracy and rendering quality compared to SLAM systems that rely on Gaussian assumptions as shown in \cref{tab:ablation_model_design}. Additionally, our system outperforms in terms of speed and parameter efficiency. Under the Gaussian assumption, depth uncertainty is typically modeled using an additional hash grid for separate estimation. This introduces extra variables that need optimization, which introduces further complexities and disturbances in the SLAM system. In \cref{fig:gaussian_vs_us}, we conducted a comparison of rendering quality and depth uncertainty between the two methods across three datasets. The superiority of our approach is evident, particularly on real-world datasets such as TUM-RGBD\cite{tumrgbd} and ScanNet\cite{scannet}, where the visualized depth uncertainty clearly highlights the advantages of our method.

\begin{figure*}[htbp]
  \centering
  \includegraphics[height=13.5cm]{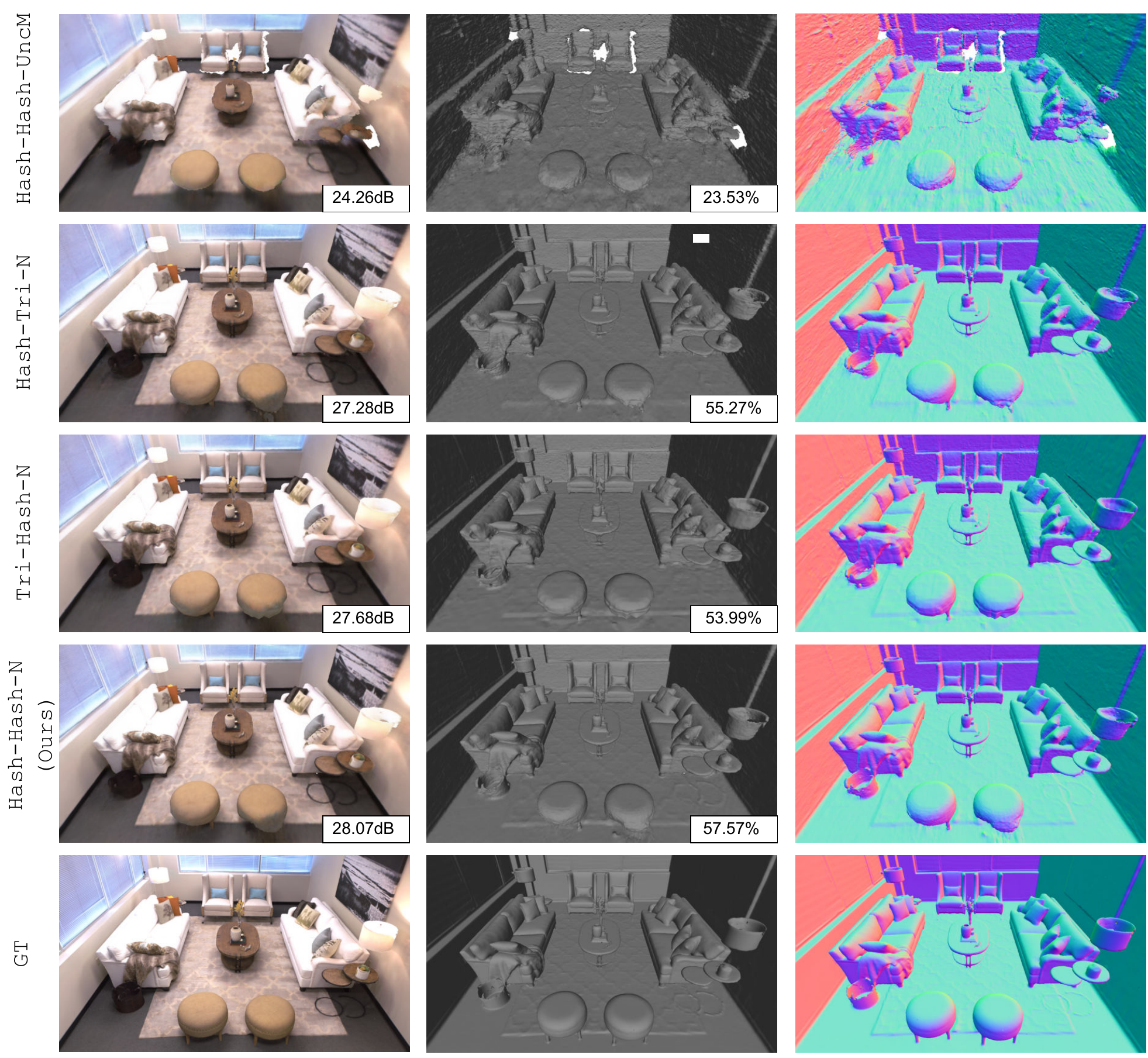}
  \caption{\textbf{Ablation on Model Design.} We compare different scene representation combinations on Replica\cite{replica} \texttt{room0} and evaluate with metrics PSNR and completion ratio$[<1cm \%]$. \texttt{Hash-Hash-UncM} denotes using hash grids for geometry and appearance, with a learnable uncertainty model. \texttt{Hash-Tri-N} uses a hash grid for geometry, a tri-plane for appearance, and our proposed model-free method for uncertainty estimation. The results show that using hash grids for both geometry and appearance, combined with the model-free uncertainty definition, achieves the best results.}
  \label{fig:Ablation_on_Model_Design}
\end{figure*}

Moreover,  in addition to aboving scene representation, we also experimented with the memory-efficient tri-plane\cite{chan2022efficient} method for encoding geometry and appearance respectively. In \cref{tab:details_ablation_study}, rows a) through d) provide quantitative results on the Replica dataset, while \cref{fig:Ablation_on_Model_Design} presents the corresponding qualitative visualizations. The results show that using two hash grids for encoding provides the best performance.

\subsection{Model Capability Analysis}
To demonstrate the high capability of our model in reconstructing quality scenes and to fairly compare the model's upper limits, we compared our method with state-of-the-art dense implicit SLAM approaches, including ESLAM \cite{eslam} and Co-SLAM \cite{coslam} on Replica dataset\cite{replica}. We standardized the mapping iterations and tracking iterations to 30, and set the number of sampling points to 5000. The results in \cref{tab:optimization_iterations} indicate that our method achieves superior performance in terms of evaluation metrics localization accuracy ATE RMSE, reconstruction accuracy, completion ratio, PSNR, and computational efficiency.

\begin{table}[htbp]
\scriptsize
\centering
\begin{tabular}{l|cccccc}
\toprule
Method  & ATE & Acc.  & Comp. Ratio  & PSNR & Time  \\
 & (cm)$\downarrow$ &  $ (cm) \downarrow$ & $[<1cm \%] \uparrow$  & (dB) $\uparrow$ & Mins $\downarrow$  \\
\midrule
ESLAM \cite{eslam} & 0.40  & 0.91 & 63.51  & 31.63   & 21.53 \\
Co-SLAM \cite{coslam}  & 0.75 & 1.07 & 57.79  & 31.77  & 11.92 \\
ours                 & \textbf{0.29}  & \textbf{0.84} & \textbf{68.35} & \textbf{32.82} & \textbf{11.17} \\
\bottomrule
\end{tabular}
\caption{Capability analysis of the effect of the number of optimization iterations during mapping and tracking on our method's reconstruction quality.}
\label{tab:optimization_iterations}
\end{table}

\subsection{Model Convergence Speed Analysis}

\begin{figure}[htbp]
  \centering
  \includegraphics[height=6cm]{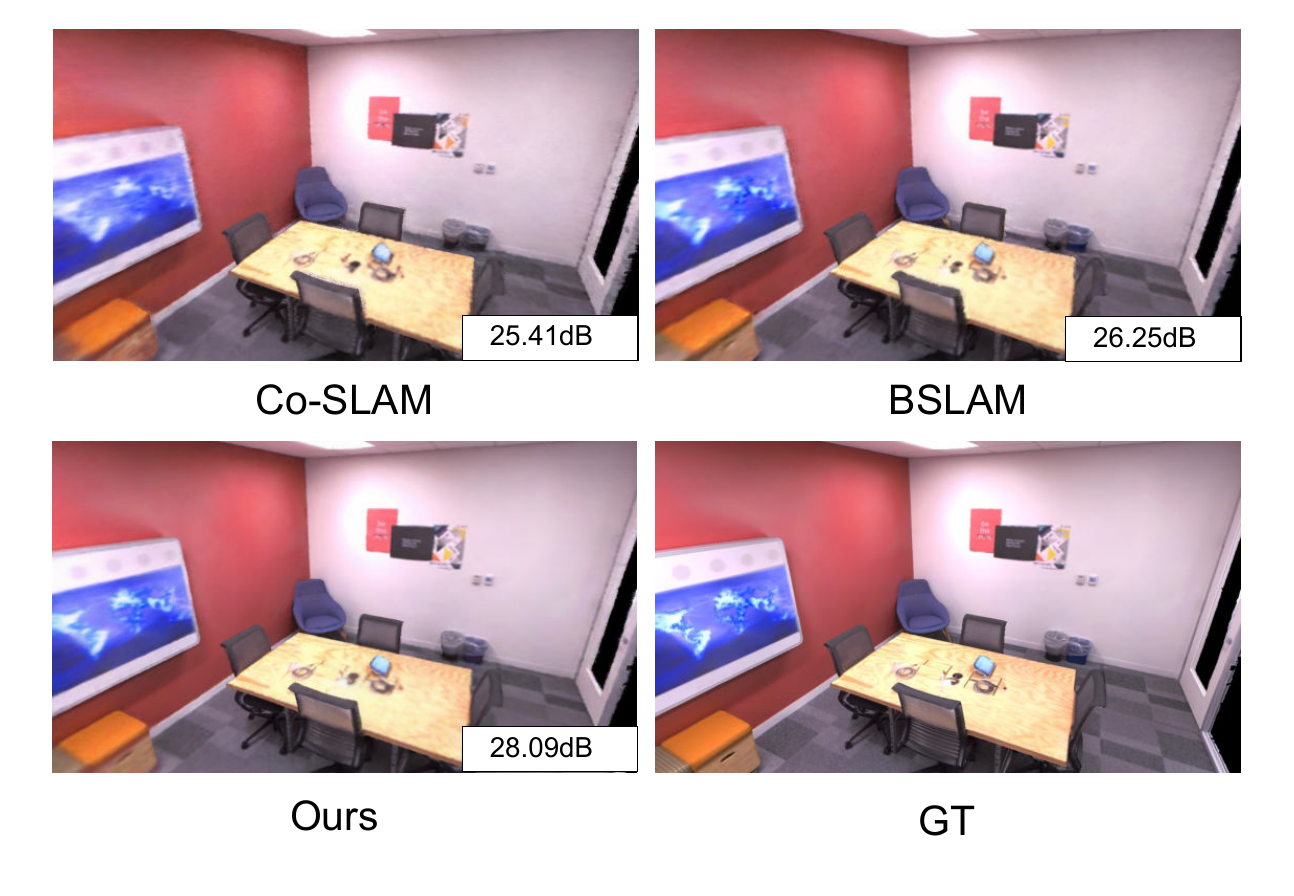}
  \caption{\textbf{Rendering Comparison on Replica dataset \cite{replica}.} Ours shows the best rendering quality compared to state-of-the-art methods BSLAM \cite{birn-slam} and Co-SLAM \cite{coslam} among dense implicit SLAM methods. Please zoom in for details.}
  \label{fig:replica_rendering}
 
\end{figure}

\begin{figure}[htbp]
   
  \centering
   \input{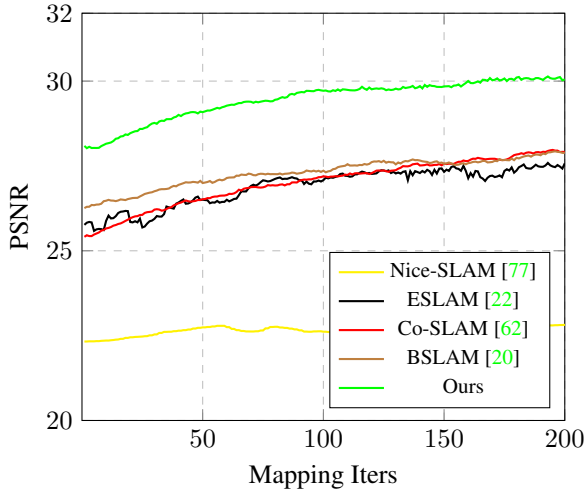}
   \caption{Comparative rendering quality convergence on the Replica dataset \cite{replica}. We set mapping iterations to 200 steps for one frame and recorded PSNR at each iteration. Our model showed stable, monotonic growth in PSNR, attributed to its decoupled scene representation. In contrast, ESLAM exhibits higher variance, and Nice-SLAM, Co-SLAM, and BSLAM have lower PSNR values, indicating slower convergence and poorer performance.}
   \label{fig:replica_fast_converge}
   \captionsetup{skip=-100pt}
\end{figure}

\begin{figure}[htbp]
  \centering
  \includegraphics[height=6cm]{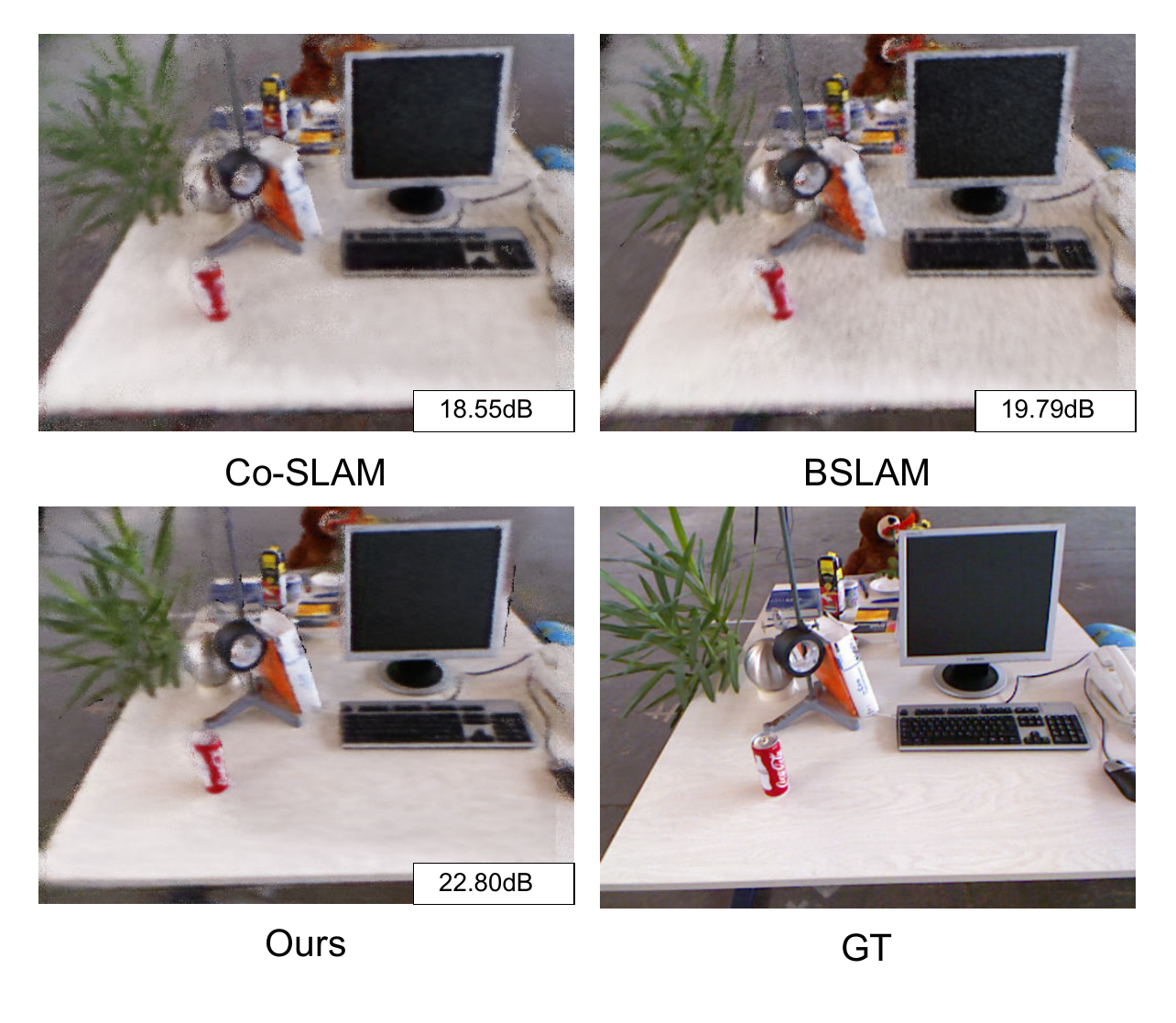}
  \caption{\textbf{Rendering Comparison on TUM RGB-D \cite{tumrgbd}.} Ours shows the best results compared to state-of-the-art methods BSLAM \cite{birn-slam} and Co-SLAM \cite{coslam} among dense implicit SLAM methods. }
  \label{fig:tumrgbd_rendering}
\end{figure}

\begin{figure}[htbp]
  \centering
   \input{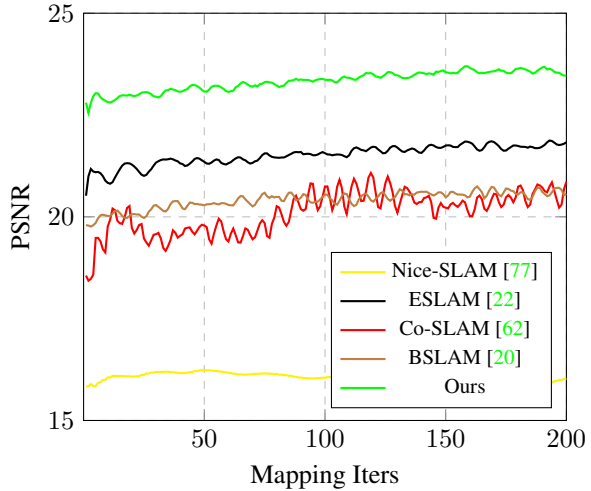}
   \caption{Comparative rendering quality convergence on TUM RGB-D \cite{tumrgbd}.  We set the mapping process iterations to 200 steps and recorded the PSNR for each iteration. The variation curve shows a stable monotonic increase, demonstrating the model's stability on real-world challenging datasets. In contrast, Co-SLAM's variation curve oscillated, reflecting poorer stability in rendering. Meanwhile, Nice-SLAM, ESLAM, and BSLAM showed suboptimal results due to insufficient model capability and slower convergence.}
   \label{fig:tum_fast_converge}
\end{figure}

To compare model convergence speed and rendering quality, we conducted experiments on the synthetic Replica dataset and the realistic TUM RGB-D dataset. \cref{fig:replica_rendering} and \cref{fig:replica_fast_converge} illustrate the qualitative rendering quality and quantitative changes over iterations on the Replica dataset. Our model exhibited the best rendering quality with a stable, monotonically increasing curve, attributed to its decoupled grid-based scene representation. On the real-world TUM RGB-D dataset, as shown in \cref{fig:tumrgbd_rendering} and \cref{fig:tum_fast_converge} our model also outperformed Nice-SLAM, ESLAM, Co-SLAM, and BSLAM. The other models showed instability (e.g., ESLAM on Replica, Co-SLAM on TUM-RGBD) and suboptimal rendering quality.

\subsection{Runtime and Memory Analysis }
In \cref{tab:runtimeMemory}, we compare runtime and memory usage, benchmarking all methods on NVIDIA GeForce RTX 4090 GPU using \texttt{room0} of Replica \cite{replica}, \texttt{scene0000} of ScanNet\cite{scannet} and \texttt{freiburg2-xyz} of TUM-RGBD\cite{tumrgbd}. We report tracking and mapping times per iteration and compare iteration steps to show convergence speed. The results show that our method achieved competitive real-time performance compared to Co-SLAM.

\begin{table}[htbp]
\vspace{-2mm}
\tiny
\centering
\begin{tabular}{ l | l | c @{\hspace{5pt}}  c @{\hspace{5pt}} c @{\hspace{5pt}} c @{\hspace{5pt}} c @{\hspace{5pt}} }
\hline 
  &                    \multirow{2}{*}{Method}    & Tracking                 & Mapping           & \multirow{2}{*}{FPS$\uparrow$} & Time & \multirow{2}{*}{params.$\downarrow$} \\ 
  &                             & [ms x it.] $\downarrow$     & [ms x it.] $\downarrow$ &  & Mins$\downarrow$                       \\        
\hline 
\multirow{5}{*}{\rotatebox[origin=c]{90}{Replica}} &

                        Nice-SLAM \cite{niceslam}      & 6.5 x 10 & 29.3 x 0  & 1.8   & 18.51 & 12.13M \\

                        &  Co-SLAM \cite{coslam}          & \textbf{4.6 x 10} & \textbf{6.6 x 10}   & \textbf{9.07}  & \textbf{3.67}  & \textbf{1.72M} \\

                        & ESLAM \cite{eslam}             & 7.9 x 8  & 18.8 x 15 & 5.55  & 6.01  & \underline{6.78M} \\

                        &  BSLAM \cite{pointslam}         & 11 x 20  & 15 x 20    & 1.66   & 20.3  & 17.38M \\

                        & Ours                           & \underline{7.0 x 8} & \underline{8.1 x 13}  & \underline{8.37}   & \underline{4.02}  & 12.69M \\
                        
\hline 
\multirow{5}{*}{\rotatebox[origin=c]{90}{ScanNet}} 

                        & Nice-SLAM \cite{niceslam}      & 11.3 x 50 & 41.2x60  & 1.34   & 57.8   & 22.04M \\

                        &  Co-SLAM \cite{coslam}          & \textbf{5.6 x 20} & \textbf{12.7 x 10}   & \textbf{5.7}  & \textbf{17.2}  & \textbf{1.74M} \\

                        & ESLAM \cite{eslam}             & 13.41 x 30  & 22.5 x 30 & 1.57  & 40.6  & 17.63M \\

                        &  BSLAM \cite{pointslam}         & 250 x 20  & 400 x 20    & 0.52   & 176  & 18.5M \\

                        & Ours                           & \underline{6.3 x 20} & \underline{11.7 x 30}  & \underline{4.88}   & \underline{20.8}  & \underline{3.39M} \\
                        
\hline 

\multirow{5}{*}{\rotatebox[origin=c]{90}{TUM RGB-D}} 

                        & Nice-SLAM \cite{niceslam}      & 33 x 200 & 103 x 60  & 0.09   &  577  & 120.95M \\

                        &  Co-SLAM \cite{coslam}          & \textbf{4.3 x 20} & \textbf{15.6 x 10}   & \textbf{6.4}  & \textbf{8.5}  & \textbf{1.68M} \\

                        & ESLAM \cite{eslam}             & 20.5 x 200  & 22.3 x 60 & 0.33  & 175  & 9.51M \\

                        &  BSLAM \cite{pointslam}         & 251 x 20  &  370 x 20    & 0.95   & 59  & 19.76M \\

                        & Ours                           & \underline{12.3 x 20} & \underline{13.7 x 20}  & \underline{2.7}   & \underline{21.3}  & \underline{3.58M} \\
\hline 
\end{tabular}
\caption{Runtime and Memory Usage Comparison.}
\label{tab:runtimeMemory}

\end{table}

\subsection{Ablation on Reweighting Term}
Here, corresponding to Section 4.3 of the main paper, we provide further explanation of the reweighting term to validate our choice. In the tracking and mapping processes, the loss functions consist of three loss terms: ($\mathcal{L}_{sdf}$, $\mathcal{L}_{dep}$, $\mathcal{L}_{rgb}$). We aim to use pixel-level uncertainty to select effective information and progressively filter out outliers to enhance localization accuracy and rendering quality. If reweighting is applied, we denote it as $Y$, and if not, we denote it as $N$. For example, $YYY-YYN$ means, we reweight all ($\mathcal{L}_{sdf}$, $\mathcal{L}_{dep}$, $\mathcal{L}_{rgb}$) three terms in tracking process, and only reweight ($\mathcal{L}_{sdf}$, $\mathcal{L}_{dep}$) in mapping process.

\begin{figure*}[htbp]
  \centering
  \includegraphics[height=7cm]{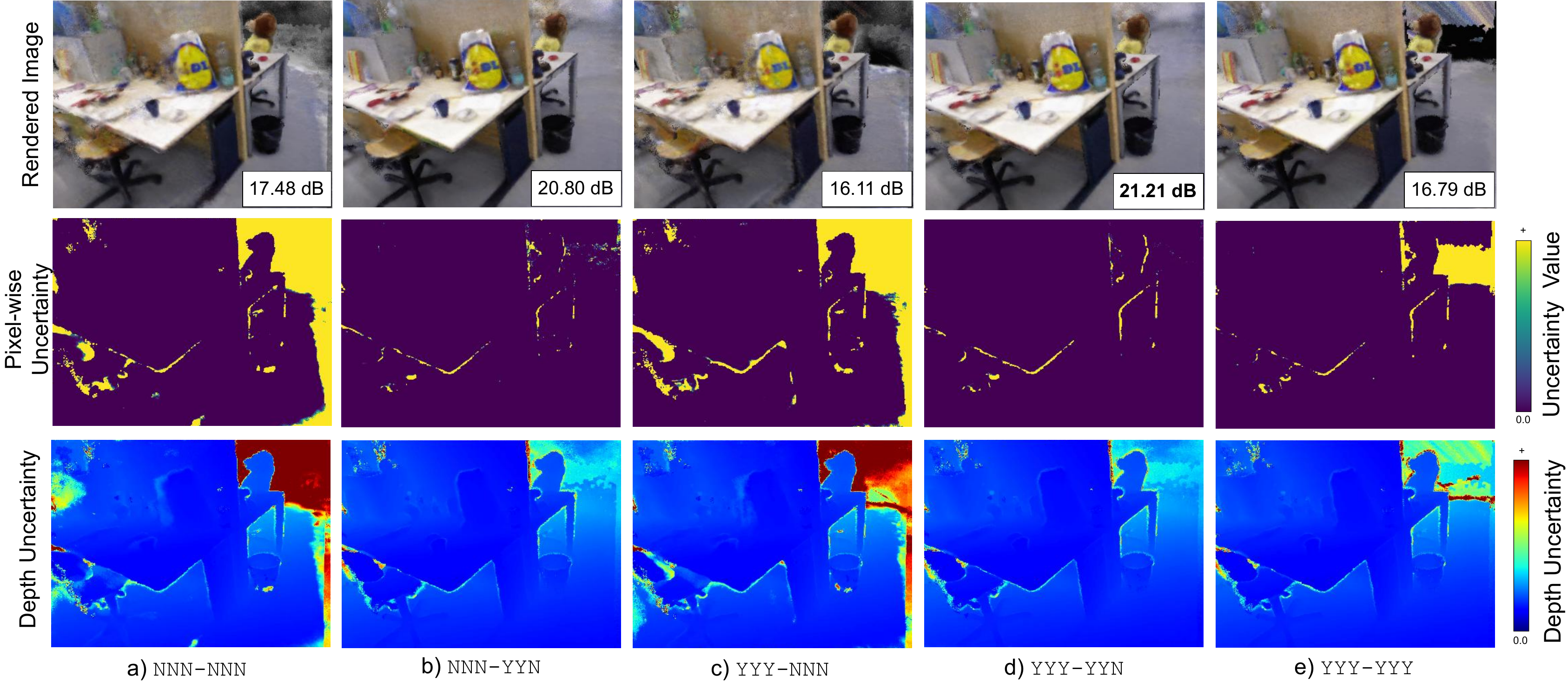}
  \caption{\textbf{Ablation on Reweighting.} In the tracking and mapping processes, the loss functions consist of three loss terms: ($\mathcal{L}_{sdf}$, $\mathcal{L}_{dep}$, $\mathcal{L}_{rgb}$). If reweighting is applied, we denote it as $Y$, and if not, we denote it as $N$. Column d) \texttt{YYY-YYN} indicates that we apply pixel-level uncertainty reweighting to all terms except for the color loss term $\mathcal{L}_{rgb}$ in the mapping process. With this uncertainty-guided reweighting strategy, we achieve the best rendering quality and depth estimation.}
  \label{fig:ablation_on_reweighting}
\end{figure*}

As shown in \cref{fig:ablation_on_reweighting}, column d) yields the optimal results. Not only does it produce the highest quality rendered color image (highest PSNR [dB]), but the pixel-level uncertainty map and the depth uncertainty map also demonstrate higher quality depth information estimation. Compared to column e), where we do not apply reweighting to the color loss term during the mapping process, our approach compensates effectively for invalid depth caused by the sensor itself, resulting in finer geometric reconstruction.

\noindent \null
\noindent \phantom{This is an invisible line of text that helps with layout.}

\begin{algorithm*}[tbp]
\noindent \phantom{This is an invisible line of text that helps with layout.}
\caption{Our Uncertainty-Aware Algorithm}
\begin{algorithmic}[1]
\State $i= 1$ \Comment{Initialize index}
\State $P$ \Comment{Estimated camera pose}
\State $n$ \Comment{Fixed-frequency for constant global BA}
\State $N$ \Comment{Number of frames of current RGB-D sequence}
\State $\theta$ \Comment{Scene representation}
\State $Optimize ()$ \Comment{Optimazation function with pixel-level uncertainty reweighting}

\While {$i < N$}
    \If {$i=1$}
        \State $P_1 = P_1^{gt}$ \Comment{Initialize first camera pose with ground truth}
        \State Optimize ($\theta_1$) \Comment{Optimize scene representation at the first frame}
        \State $i = i+1 $
    \EndIf

    \If {$i > 1$}
        \State Optimize ($P_i$)  \Comment{Tracking process for each frame}
    \EndIf

    \If {$\beta > \beta_{unc}$} \Comment{Uncertainty check}

    \State Optimize ($\theta_{local}, P_{local}$) \Comment{Local BA}
    \State $i = i+1 $
    
    \ElsIf {$OC_{cov} > \tau_{cov
} $} \Comment{Co-Visibility check}

    \State Optimize ($\theta_{LLCO}, P_{LLCO}$)\Comment{Local loop closure optimization}
    \State $i = i+1 $
    \EndIf
    
    \If {$i \bmod n == 0$}
        \State Optimize ($\theta_{global}, P_{global}$) \Comment{Global BA for every $n$ frame}
    \State $i = i+1 $
    \EndIf    

\EndWhile
\end{algorithmic}
\end{algorithm*}

\section{Per-Scene Breakdown of the Results.}
\label{sec:per_scene_resuls}
\begin{figure*}[t]
  \centering
  \includegraphics[height=20.5cm]{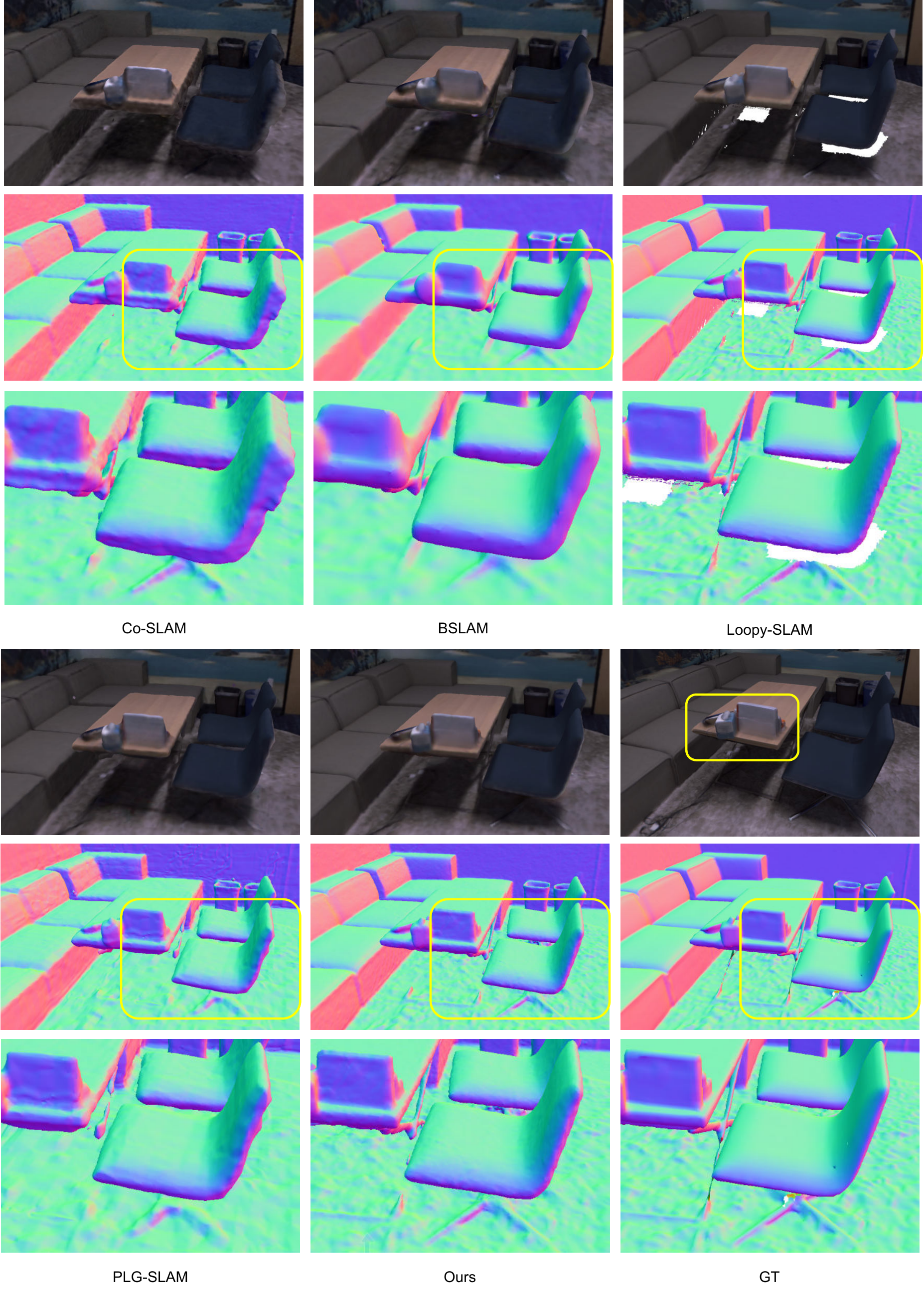}
  \caption{\textbf{Mesh Evaluation on Replica\cite{replica}} \texttt{Office-0}. Notably, our method can present fine geometric structures while also achieving better scene completion for unobserved regions compared to explicit Loopy-SLAM \cite{loopyslam}. Compared to implicit methods such as Co-SLAM \cite{coslam}, BSLAM \cite{birn-slam}, and PLG-SLAM\cite{plgslam}, our method captures finer high-frequency geometric details. For example, the chair back, chair legs, and the carpet. For appearance, rendered objects on the table are also better.} 
  \label{fig:office0}
\end{figure*}

\begin{figure*}[t]
  \centering
  \includegraphics[height=21cm]{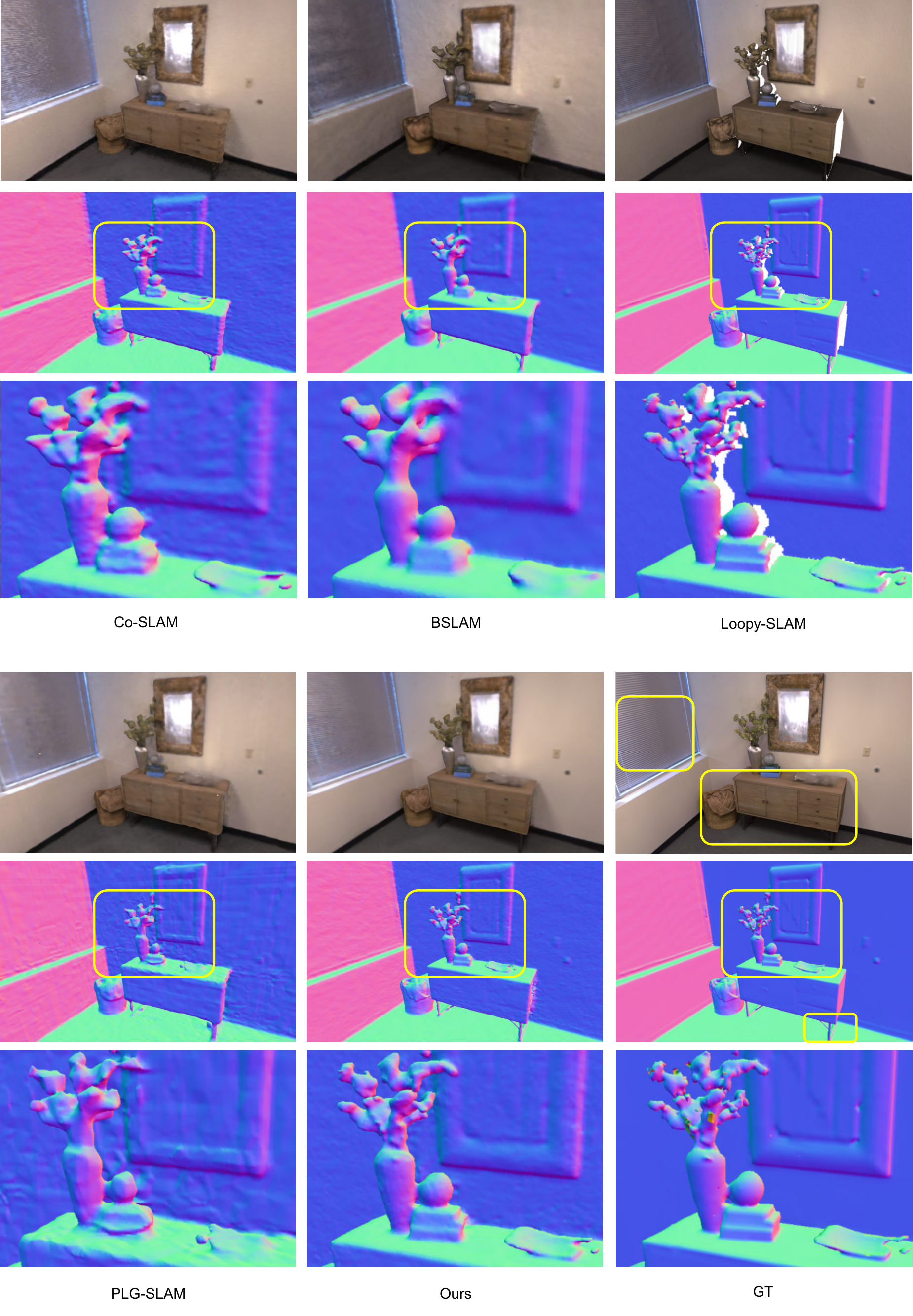}
  \caption{\textbf{Mesh Evaluation on Replica \cite{replica}} \texttt{Room-2}. Our method achieves finer geometric and appearance reconstruction. For appearance: the patterns on the curtains and the detailed textures on the cabinet surface. For geometry: the vase on the cabinet and the cabinet legs. Please zoom in for more details. }
  \label{fig:room2}
\end{figure*}

\begin{figure*}[t]
  \centering
  \includegraphics[height=21cm]{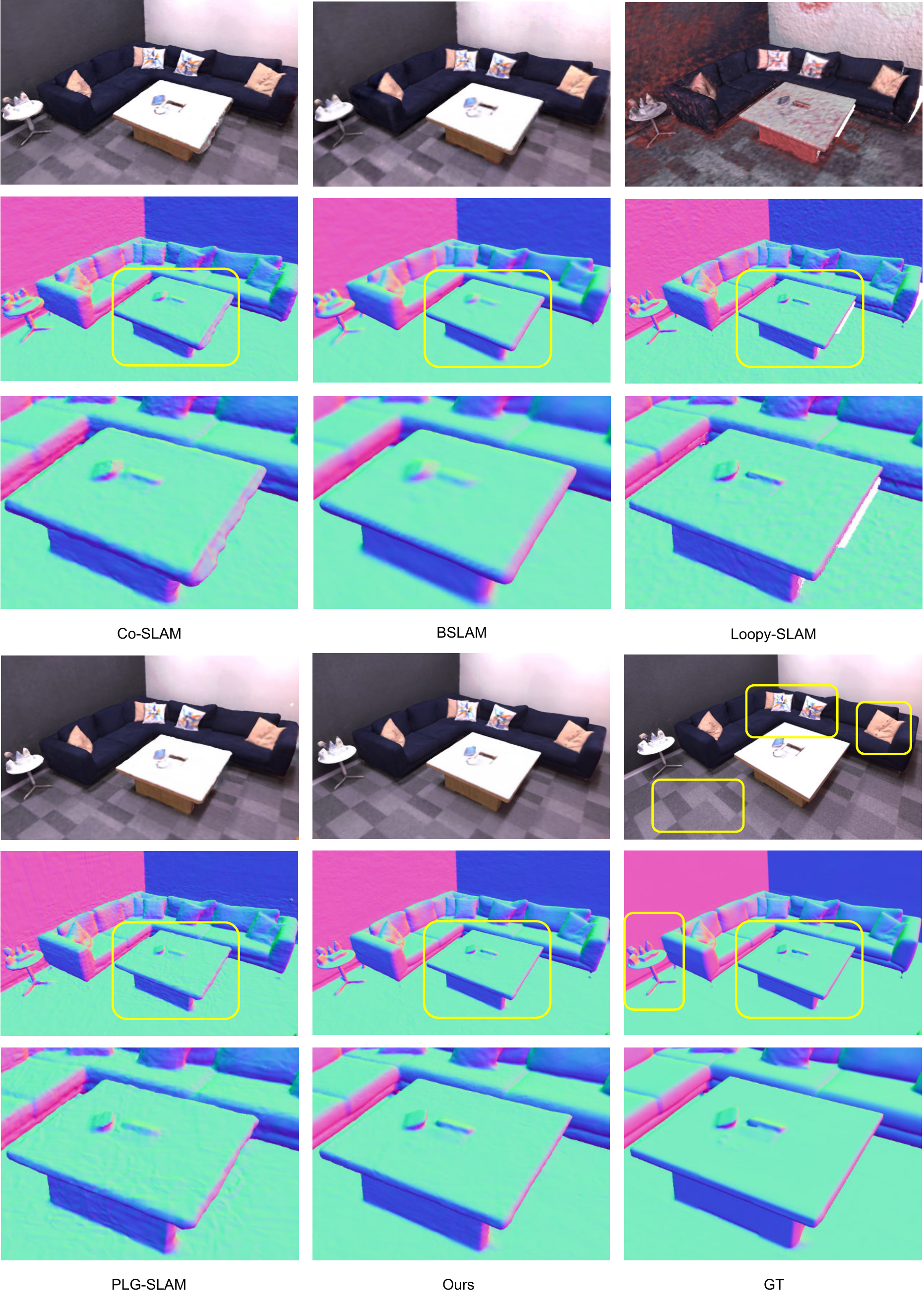}
  \caption{\textbf{Mesh Evaluation on Replica \cite{replica}} \texttt{Office-2}. For appearance: our rendered floor has higher quality, clearly distinguishing the floor patterns, as well as the textures on the pillows on the sofa. For geometry: we zoomed in on the table, and our method reconstructs sharper edges and smoother surfaces. }
  \label{fig:office2}
\end{figure*}

\begin{figure*}[t]
  \centering
  \includegraphics[height=21cm]{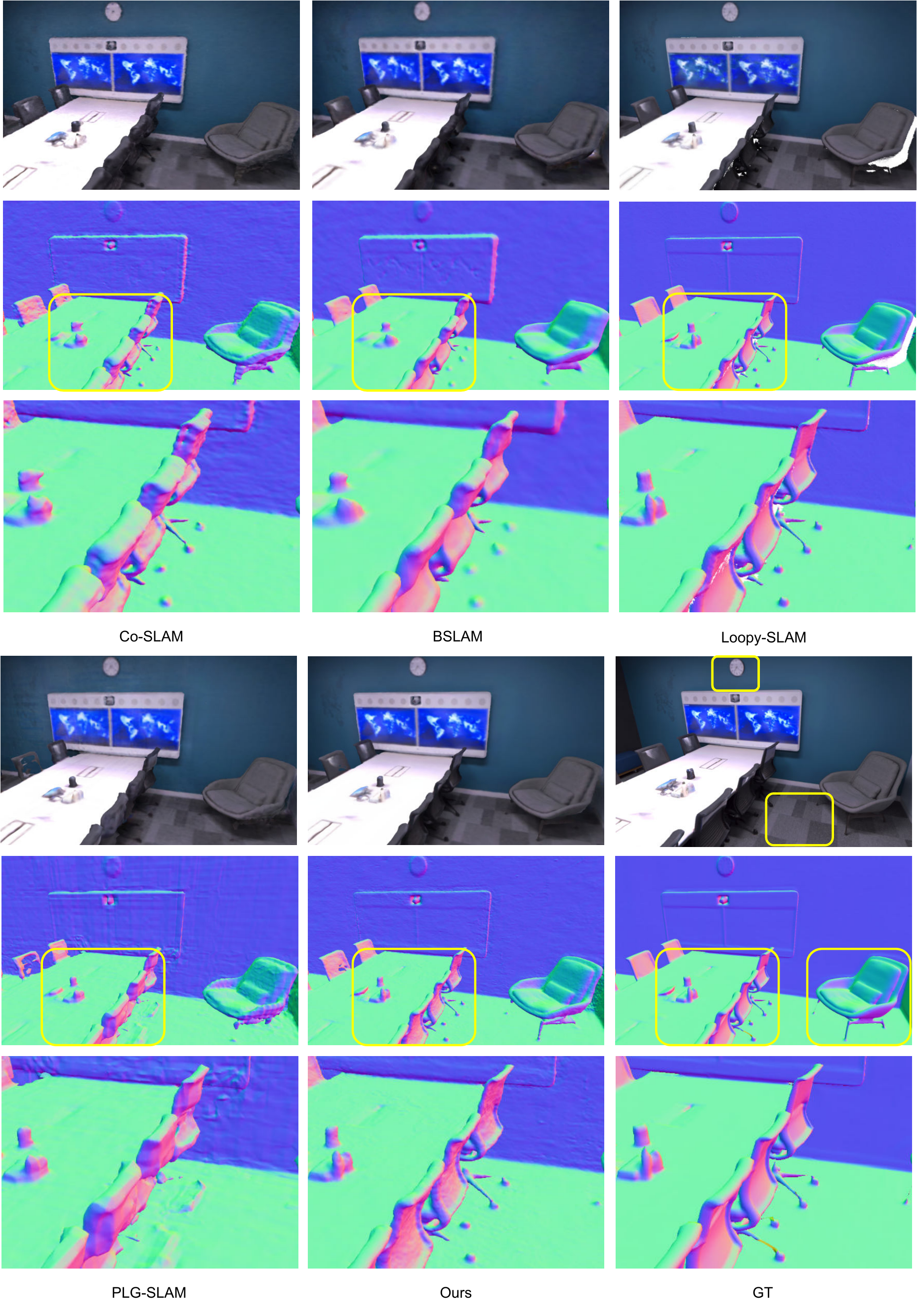}
  \caption{\textbf{Mesh Evaluation on Replica \cite{replica}} \texttt{Office-4}. For appearance: our rendered floor quality is higher, clearly distinguishing the floor patterns, as well as the clock on the wall. For geometry: we reconstructed more of the office chair's geometric structure, such as the legs and the backrest. The geometry of the sofa in the corner also demonstrates the superiority of our algorithm. }
  \label{fig:office4}
\end{figure*}

\begin{figure*}[t]
  \centering
  \includegraphics[height=20.5cm]{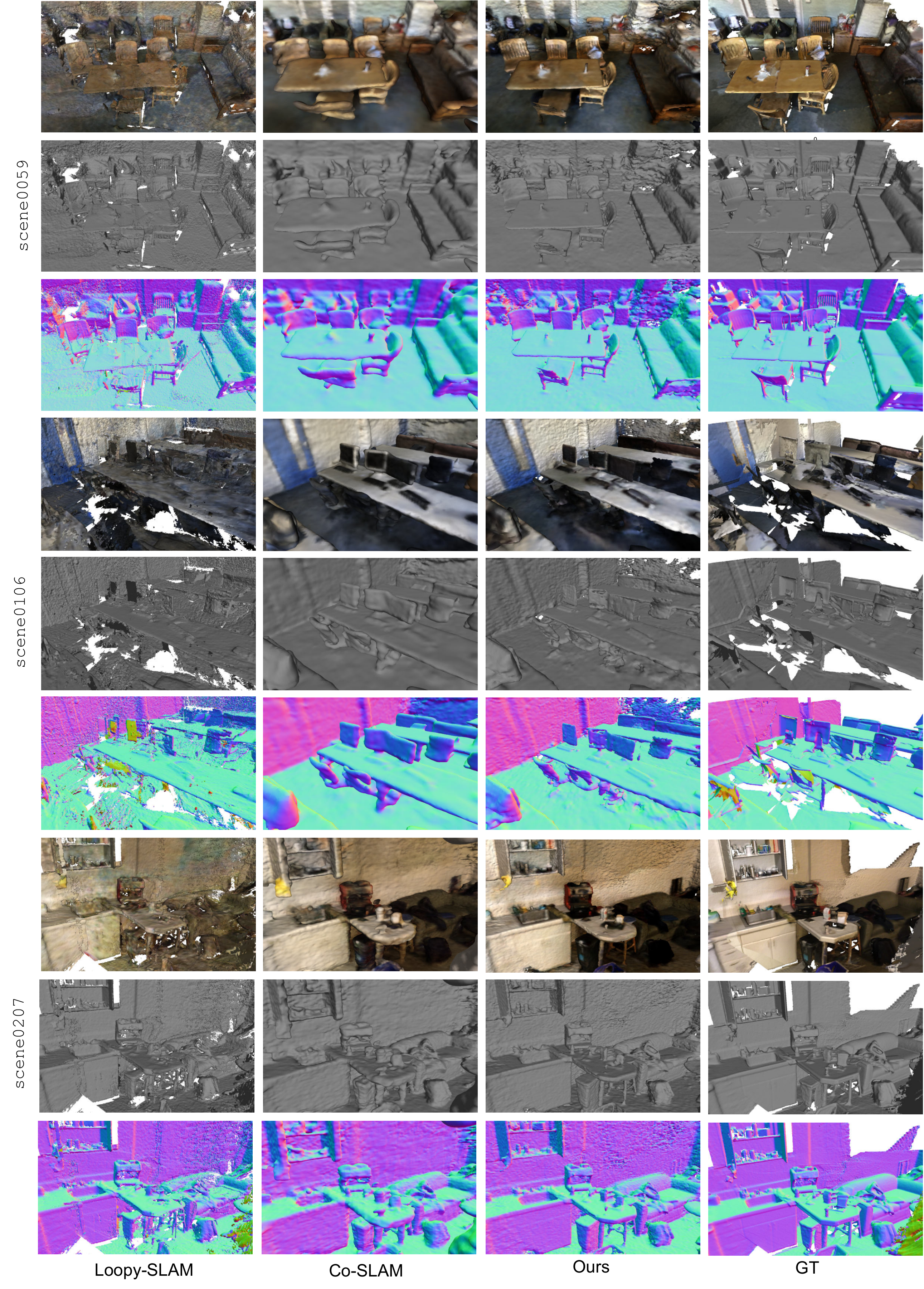}
  \caption{\textbf{Mesh Evaluation on ScanNet\cite{scannet}.} Explicit Loopy-SLAM\cite{loopyslam} and implicit Co-SLAM\cite{coslam} are listed here for comparison. For appearance: Our method achieves higher rendering quality compared to the ground truth (GT) mesh, as seen texture of chairs, objects on the desk in \texttt{scene0059}, and the coffee machine on the table in \texttt{scene0207}. For geometry: More detailed and complete results are reconstructed, such as table and chairs in \texttt{scene0059}, the surface of desks in \texttt{scene0207}.}
  \label{fig:sup_scannet}
\end{figure*}

\begin{figure*}[t]
  \centering
  \includegraphics[height=19cm]{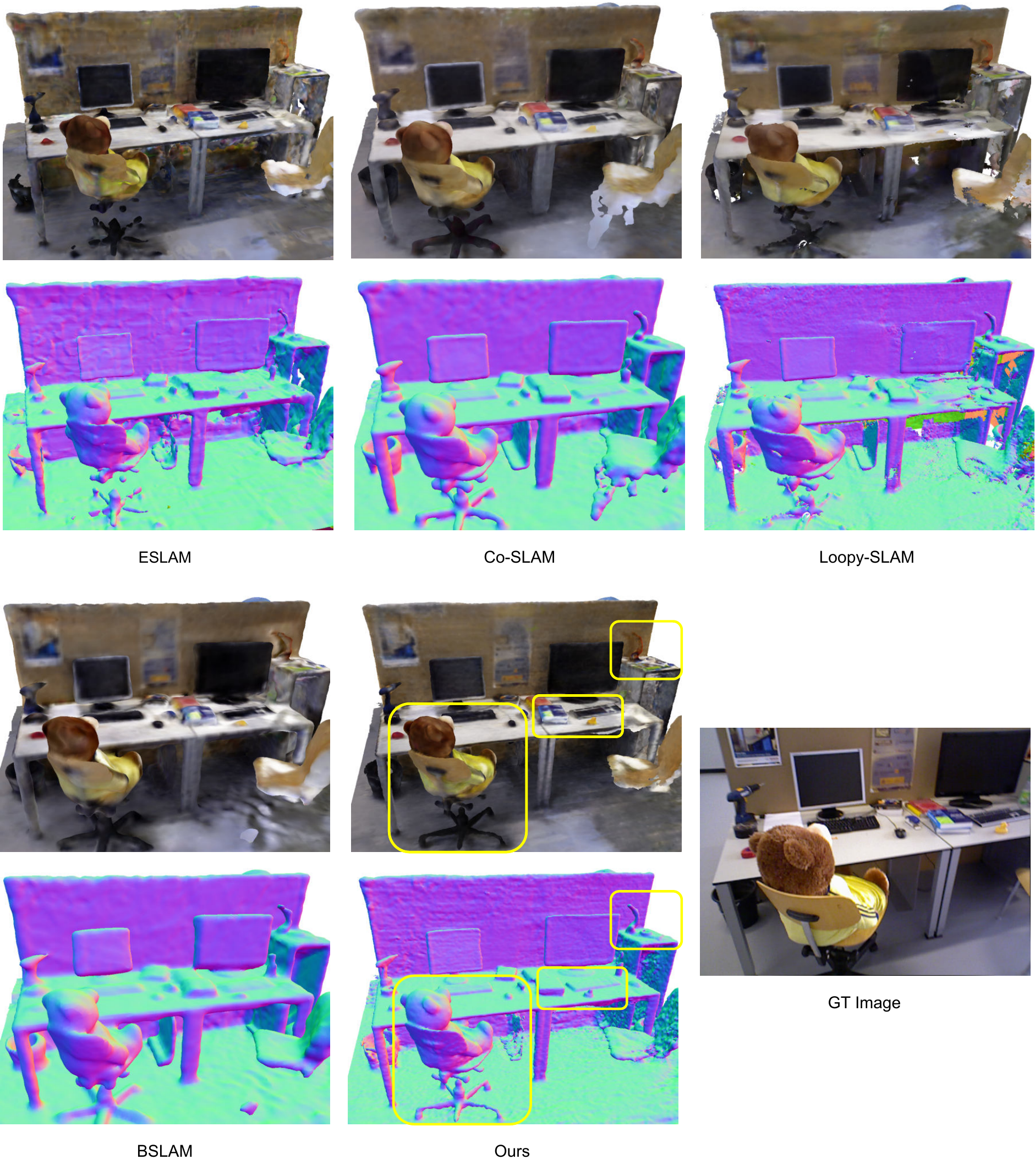}
  \caption{\textbf{Mesh Evaluation on TUM RGB-D\cite{tumrgbd}.} Because there is no ground truth mesh for the TUM RGB-D dataset, we provide an image to facilitate qualitative comparison. We extensively compare the reconstruction quality with implicit methods such as ESLAM \cite{eslam}, Co-SLAM \cite{coslam}, and BSLAM \cite{birn-slam}, as well as the explicit method Loopy-SLAM \cite{loopyslam}. The results show that our method achieves superior quality in both rendering and geometry. Our method captures finer geometric details and higher fidelity rendering, such as the legs of the chair, the teddy bear, and the captured objects on the table.}
  \label{fig:office-view-1}
\end{figure*}

\begin{figure*}[t]
  \centering
  \includegraphics[height=20cm]{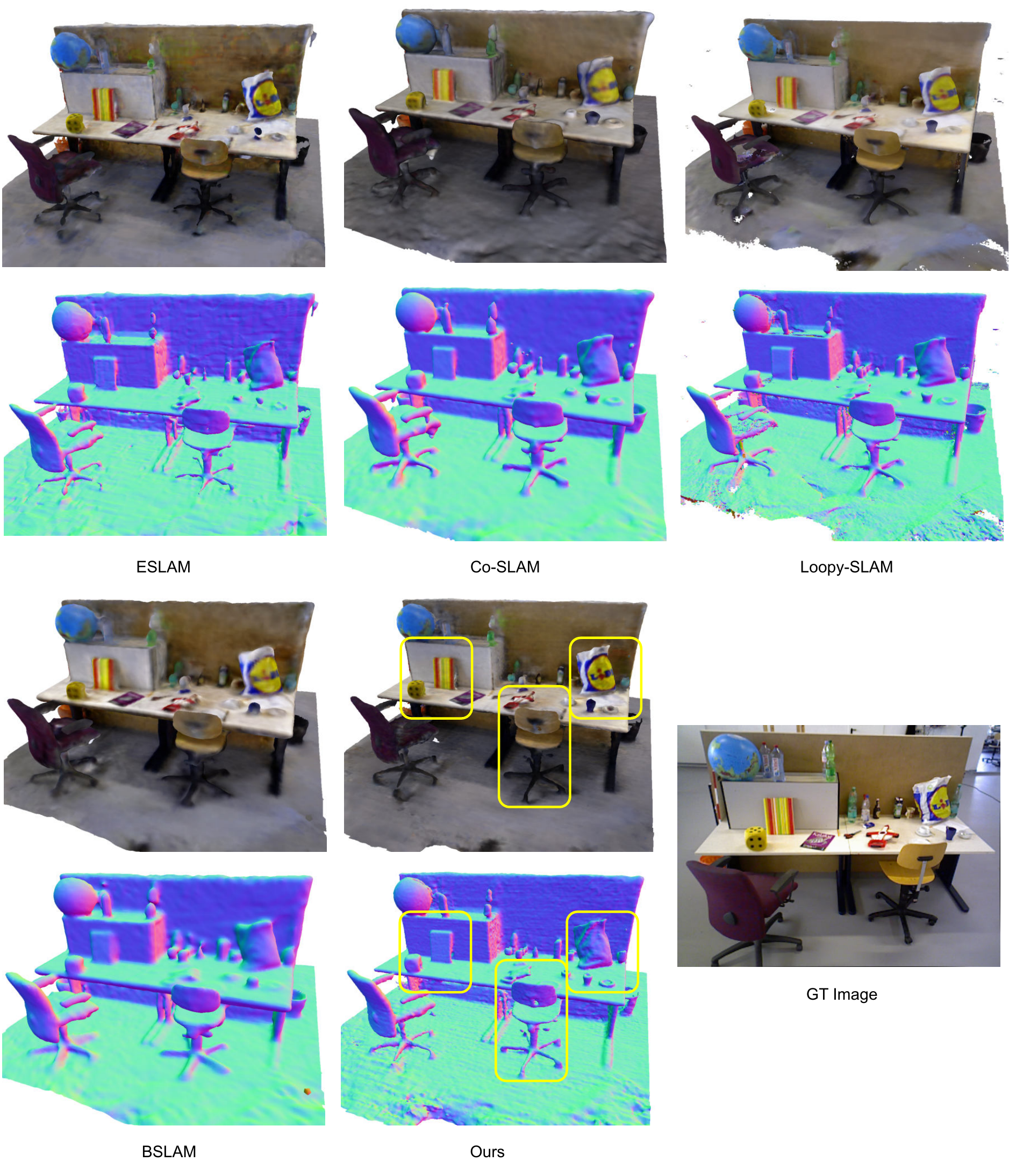}
  \caption{\textbf{Mesh Evaluation on TUM RGB-D\cite{tumrgbd}.} Because there is no ground truth mesh for the TUM RGB-D dataset, we provide an image to facilitate qualitative comparison. For example, our method accurately reconstructs details such as the Rubik's cube on the table, the shopping bag, and the chair.}
  \label{fig:office-view-2}
\end{figure*}

\begin{figure*}[t]
  \centering
  \includegraphics[height=15cm]{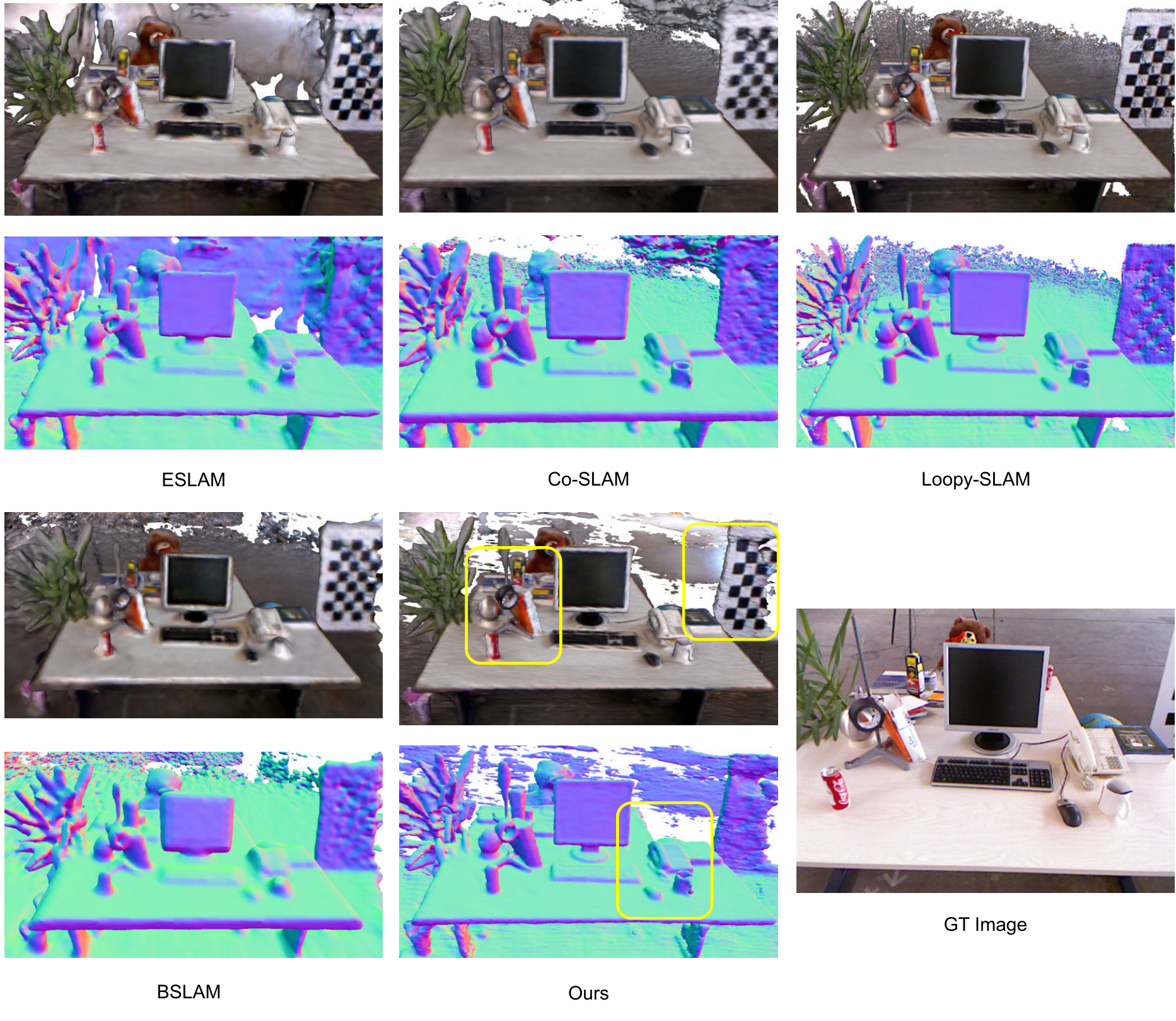}
  \caption{\textbf{Mesh Evaluation on TUM RGB-D\cite{tumrgbd}.} While ESLAM\cite{eslam} and BSLAM\cite{birn-slam} can not capture geometric details such as cup and mouse on table, Co-SLAM\cite{coslam} can not reconstruct thin geometric structure, such thin table surface. Our method shows outstanding performance.}
  \label{fig:office-xyz}
\end{figure*}

In this section, we provide more per-scene qualitative and quantitative results. \cref{tab:per_scene_replica_3d_evaluation} and \cref{tab:per_scene_replica_2d_evaluation} present the quantitative results for 3D and 2D metrics on the Replica dataset \cite{replica} for each scene, respectively. \cref{fig:office0,fig:office2,fig:office4,fig:room2} show the qualitative reconstructed meshes. The results for Nice-SLAM\cite{niceslam}, Co-SLAM \cite{coslam}, ESLAM \cite{eslam}, and BSLAM \cite{birn-slam} are obtained using their open-source code over five experimental runs. For PLG-SLAM\cite{plgslam}, the authors only provide us the reconstructed meshes on the Replica dataset, so the qualitative comparison is not provided here. Additionally, although this paper primarily investigates the application of uncertainty in real-time implicit NeRF-SLAM, for a broader qualitative comparison of reconstruction quality, we also include explicit scene representations, such as Loopy-SLAM \cite{loopyslam}. Overall, the results demonstrate that our method achieves finer reconstructions among all implicit methods while addressing the hole-filling limitations of explicit scene representations. For real-world datasets, \cref{fig:sup_scannet} shows the reconstruction results on ScanNet \cite{scannet}, and \cref{fig:office-view-1,fig:office-view-2,fig:office-xyz} display our reconstruction results on TUM RGB-D \cite{tumrgbd}. These results indicate that our method achieves more precise detail reconstruction and high-fidelity rendering, which we attribute to robust scene representation and an uncertainty-aware strategy.

\begin{table*}[ht]
\centering
\begin{tabular}{lcccc|c}
\toprule
\textbf{Methods} & \multicolumn{4}{c}{\textbf{Reconstruction \& Rendering}} & \multicolumn{1}{c}{\textbf{Localization [cm]}} \\
 & Acc. & Comp. Ratio & Depth L1 & PSNR  & RMSE \\
\midrule
a) Gaussian assumption uncertainty with third grid & 1.79 & 31.52 & 3.75 & 27.33 & 1.51 \\
b) Coupled scene representation with one grid & 1.05 & 63.15 & 0.94 & 30.12 & 0.51 \\
c) Grid for geometry and tri-plane for appearance& 1.01 & 64.69 & 0.93 & 30.98& 0.47 \\
d) Tri-plane for geometry and grid for appearance& 1.17 & 63.82 & 0.97 & 21.32 & 0.50\\

\hline
e) w/o camera pose optimization in mapping & 1.89 & 26.88 & 1.76 & 27.56 & 3.52 \\
f) Only global BA in mapping    & 0.96 & 66.01 & 0.95 & 31.32 & 0.49 \\
g) Only local BA in mapping     & 1.01 & 65.21 & 0.91 & 30.87 & 0.55 \\
h) Global + local BA in mapping & 0.94 & 66.34 & 0.89 & 31.51 & 0.45 \\

Ours & \textbf{0.92} & \textbf{66.86} & \textbf{0.89} & \textbf{31.62} & \textbf{0.45} \\
\bottomrule

\end{tabular}
\caption{We conduct experiments on Replica \cite{replica} to verify the effectiveness of our method. Our full model achieves better completion reconstructions and more accurate pose estimation results.}
\label{tab:details_ablation_study}
\end{table*}

\begin{table*}[htbp]
\centering
\renewcommand{\arraystretch}{1.3}
\begin{tabular}{l @{\hspace{10pt}} lccccccccc}
\hline & & room0 & room1 & room2 & office0 & office1 & office2 & office3 & office4 & Avg. \\
\hline 
\multirow{6}{*}{\rotatebox[origin=c]{90}{Nice-SLAM]\cite{niceslam}}} & Depth L1 $[\mathrm{cm}] \downarrow$    
                                              & 2.51 & 2.65 & 3.37 & 2.12 & 2.20 & 4.53 & 4.30 & 3.79 & 3.18 \\
& Acc. $[\mathrm{cm}] \downarrow$             & 1.51 & 1.44 & 1.62 & 1.34 & 1.02 & 1.71 & 2.02 & 4.55 & 1.90 \\
& Comp. $[\mathrm{cm}] \downarrow$            & 1.50 & 1.39 & 1.54 & 1.42 & 1.08 & 1.57 & 1.82 & 1.94 & 1.53 \\
& Comp. Ratio $[<5cm \%] \uparrow$ & 98.33 & 98.81 & 97.37 & 97.6 & 98.08 & 97.65 & 95.81 & 95.92 & 97.45 \\
& Comp. Ratio $[<3cm \%] \uparrow$ & 95.20 & 95.30 & 91.45 & 94.82 & 95.52 & 92.91 & 90.30 & 88.10 & 92.95 \\
& Comp. Ratio $[<1cm \%] \uparrow$ & 32.63 & 39.07 & 35.17 & 42.37 & 67.39 & 31.22 & 24.07 & 23.48 & 36.93 \\

\hline 
\multirow{6}{*}{\rotatebox[origin=c]{90}{Co-SLAM\cite{coslam}}}  & Depth L1 $[\mathrm{cm}] \downarrow$ 
                                              & 1.51 & 2.38 & 3.00 & 1.51 & 1.46 & 2.68 & 2.81 & 1.85 & 2.15 \\
& Acc. $[\mathrm{cm}] \downarrow$             & 1.11 & 1.33 & 1.22 & 0.99 & 0.71 & 1.36 & 1.29 & 1.24 & 1.16 \\
& Comp. $[\mathrm{cm}] \downarrow$            & 1.04 & 1.30 & 1.18 & 0.90 & 0.71 & 1.29 & 1.35 & 1.15 & 1.12 \\
& Comp. Ratio $[<5cm \%] \uparrow$ & 98.84 & 99.05 & 97.85 & 98.52 & 98.62 & 97.52 & 98.65 & 97.12 & 98.27 \\
& Comp. Ratio $[<3cm \%] \uparrow$ & 97.82 & 97.15 & 94.45 & 97.87 & 97.57 & 96.28 & 95.89 & 94.46 & 96.44 \\
& Comp. Ratio $[<1cm \%] \uparrow$ & 54.69 & 40.08 & 55.47 & 71.35 & 87.41 & 46.93 & 39.21 & 52.35 & 55.94 \\

\hline 
\multirow{6}{*}{\rotatebox[origin=c]{90}{ESLAM\cite{eslam}}} & Depth L1 $[\mathrm{cm}] \downarrow$ 
                                              & 0.97 & 1.07 & 1.28 & 0.86 & 1.26 & 1.71 & 1.43 & 1.06 & 1.18 \\
& Acc. $[\mathrm{cm}] \downarrow$             & 1.07 & 0.85 & 0.93 & 0.85 & 0.83 & 1.02 & 1.21 & 1.15 & 0.97 \\
& Comp. $[\mathrm{cm}] \downarrow$            & 1.12 & 0.88 & 1.05 & 0.96 & 0.81 & 1.09 & 1.42 & 1.27 & 1.05  \\
& Comp. Ratio $[<5cm \%] \uparrow$ & 99.06 & 99.64 & 98.84 & 98.34 & 98.85 & 98.60 & 96.80 & 97.65 & 98.47\\
& Comp. Ratio $[<3cm \%] \uparrow$ & 98.84 & 99.24 & 96.73 & 97.89 & 98.02 & 98.02 & 96.31 & 96.54 & 97.70 \\
& Comp. Ratio $[<1cm \%] \uparrow$ & 53.06 & 70.27 & 62.15 & 73.11 & 84.13 & 59.32 & 46.93 & 49.06 &  62.25 \\

\hline 
\multirow{6}{*}{\rotatebox[origin=c]{90}{BSLAM\cite{birn-slam}}} & Depth L1 $[\mathrm{cm}] \downarrow$ 
                                              & 1.44 & 1.43 & 3.05 & 1.64   & 1.95 & 4.18 & 4.10 & 2.43 & 2.52 \\
& Acc. $[\mathrm{cm}] \downarrow$             & 1.02 & 0.92 & 1.01 & 0.86   & 0.69 & 1.46 & 1.75 & 1.27 & 1.12 \\
& Comp. $[\mathrm{cm}] \downarrow$            & 1.05 & 0.94 & 1.15 & 0.91   & 0.76 & 1.34 & 1.39 & 1.26 & 1.1 \\
& Comp. Ratio $[<5cm \%] \uparrow$           & 99.48 & 99.69 & 98.22 & 98.97 & 99.27 & 98.8 & 98.28 & 99.28 & 98.99\\
& Comp. Ratio $[<3cm \%] \uparrow$           & 98.53 & 98.74 & 94.98 & 97.64 & 97.68 & 94.46  & 95.57 & 97.71 & 96.91\\
& Comp. Ratio $[<1cm \%] \uparrow$           & 54.64 & 65.52 & 56.17 & 71.43 & 84.26 &  46.52  & 39.97 & 40.61 & 57.18  \\

\hline 
\multirow{6}{*}{\rotatebox[origin=c]{90}{Ours}} & Depth L1 $[\mathrm{cm}] \downarrow$ 
                                              & \textbf{0.81} & \textbf{0.77} & \textbf{1.13} & \textbf{0.70} & \textbf{1.11} & \textbf{1.52} & \textbf{1.15} & \textbf{0.99} & \textbf{0.89} \\
& Acc. $[\mathrm{cm}] \downarrow$             & \textbf{0.97} & \textbf{0.78} & \textbf{0.85} & \textbf{0.76} & \textbf{0.62}  & \textbf{0.92} & \textbf{1.10} & \textbf{1.15} & \textbf{0.92} \\
& Comp. $[\mathrm{cm}] \downarrow$            & \textbf{0.99} & \textbf{0.78} & \textbf{0.93} & \textbf{0.77} & \textbf{0.67} & \textbf{0.93} & \textbf{1.18} & \textbf{1.13} & \textbf{0.92} \\
& Comp. Ratio $[<5cm \%] \uparrow$ & \textbf{99.69} & \textbf{99.84} & \textbf{99.21} & \textbf{99.21} & \textbf{99.25} & \textbf{99.19} & \textbf{98.25} & \textbf{98.99} & \textbf{99.20} \\
& Comp. Ratio $[<3cm \%] \uparrow$ & \textbf{99.15} & \textbf{99.47} & \textbf{96.75} & \textbf{98.96} & \textbf{98.15} & \textbf{97.75} & \textbf{97.12} & \textbf{96.75} & \textbf{98.01} \\
& Comp. Ratio $[<1cm \%] \uparrow$ & \textbf{57.57} & \textbf{74.99} & \textbf{69.53} & \textbf{76.76} & \textbf{88.18} & \textbf{62.78} & \textbf{50.91} & \textbf{54.19} & \textbf{66.86} \\   

\hline
\end{tabular}  
\caption{Per-scene quantitative reconstruction evaluation on Replica \cite{replica}  dataset. Our method achieves consistently better reconstruction in comparison to Nice-SLAM \cite{niceslam}, Co-SLAM \cite{coslam}, ESLAM \cite{eslam} and BSLAM \cite{birn-slam}. We report Depth L1, reconstruction accuracy, completion, and completion ratios of 5cm, 3cm and 1cm respectively, reflecting our advantages in reconstruction geometry in detail. }
\label{tab:per_scene_replica_3d_evaluation}
\end{table*}

\begin{table*}[htbp]
\centering
\renewcommand{\arraystretch}{1.3}
\begin{tabular}{l l @{\hspace{5pt}} c @{\hspace{5pt}} c @{\hspace{5pt}} c @{\hspace{5pt}} c @{\hspace{5pt}}  c @{\hspace{5pt}} c @{\hspace{5pt}}c c @{\hspace{5pt}} c @{\hspace{5pt}} c @{\hspace{5pt}} c @{\hspace{5pt}} c}
\hline {Method} & Metric & Rm 0 & Rm 1 & Rm 2 & Off 0 & Off 1 & Off 2 & Off 3 & Off 4 & Avg. \\
\hline 
\hline \multirow{3}{*}{Nice-SLAM\cite{niceslam}}  & PSNR $[\mathrm{dB}] \uparrow$ & 22.12 & 22.47 & 24.52 & 29.07 & 30.34 & 19.66 & 22.23 & 24.94 & 24.42 \\
           & $\operatorname{SSIM} \uparrow$ & 0.689 & 0.757 & 0.814 & 0.874 & 0.886 & 0.797 & 0.801 & 0.856 & 0.809 \\
          & LPIPS $\downarrow$ & 0.330 & 0.271 & 0.208 & 0.229 & 0.181 & 0.235 & 0.209 & 0.198 & 0.233 \\
\hline \multirow{3}{*}{Vox-Fusion\cite{voxfusion}} & PSNR $[\mathrm{dB}] \uparrow$ & 22.9 & 22.36 & 23.91 & 27.79 & 29.83 & 20.33 & 23.47 & 25.21 & 24.4 \\
            & $\operatorname{SSIM} \uparrow$ & 0.683 & 0.751 & 0.798 & 0.857 & 0.876 & 0.794 & 0.803 & 0.847 & 0.801 \\
            & LPIPS $\downarrow$ & 0.303 & 0.269 & 0.234 & 0.241 & 0.184 & 0.243 & 0.213 & 0.199 & 0.236 \\

\hline \multirow{3}{*}{Co-SLAM\cite{coslam}} & PSNR $[\mathrm{dB}] \uparrow$ & 27.12 & 27.94 & 29.27  & 34.13 & 35.04 & 28.53 & 28.81 & 31.29 & 30.27 \\
            & $\operatorname{SSIM} \uparrow$   & 0.908& 0.900 & 0.935 & 0.962 & 0.970 & 0.939 & 0.942 & 0.957 & 0.939 \\
            & LPIPS $\downarrow$               & 0.316& 0.293 & 0.258 & 0.207 & 0.191 & 0.257 & 0.222 & 0.227 & 0.246 \\

\hline \multirow{3}{*}{ESLAM\cite{eslam}}& PSNR $[\mathrm{dB}] \uparrow$ & 27.10  & 28.41  & 29.16   & 34.59 & 34.29  & 28.97  & 28.57   & 30.51 & 30.19 \\
            & $\operatorname{SSIM} \uparrow$ & 0.914 & 0.910 & 0.938 & 0.966 & 0.963 & 0.946 & 0.948  & 0.948 & 0.942 \\
            & LPIPS $\downarrow$             & 0.295 & 0.294 & 0.240 & 0.178 & 0.208 & 0.239 & 0.194  & 0.295 & 0.243 \\

\hline \multirow{3}{*}{BSLAM\cite{birn-slam}}& PSNR $[\mathrm{dB}] \uparrow$ & 26.43  & 28.67  & 28.44   & 33.27 & 33.92   & 27.68  & 28.14  & 29.85   & 29.55 \\
            & $\operatorname{SSIM} \uparrow$                                 & 0.902 & 0.9179 & 0.919    & 0.950 & 0.963   & 0.933  & 0.939  & 0.9438  & 0.9335 \\
            & LPIPS $\downarrow$                                             & 0.300 & 0.2523 & 0.2618   & 0.201 & 0.195   & 0.246  & 0.205  & 0.2274  & 0.2361\\

\hline 
\multirow{3}{*}{Ours} & PSNR $[\mathrm{dB}] \uparrow$  & \textbf{28.07}  & \textbf{30.16}  & \textbf{30.87}  & \textbf{36.35} & \textbf{35.62}  & \textbf{29.98}  & \textbf{30.06}  & \textbf{31.85} & \textbf{31.62} \\
     & $\operatorname{SSIM} \uparrow$ & \textbf{0.927} & \textbf{0.940} & \textbf{0.955} & \textbf{0.978}& \textbf{0.977} & \textbf{0.961} & \textbf{0.962} & \textbf{0.965} & \textbf{0.958} \\
     & LPIPS $\downarrow$             & \textbf{0.241} & \textbf{0.201} & \textbf{0.172} & \textbf{0.145} & \textbf{0.167} & \textbf{0.231} & \textbf{0.156} & \textbf{0.169} & \textbf{0.185} \\

\hline
\end{tabular}
\caption{Per-scene quantitative rendering evaluation on Replica \cite{replica}. Our method achieves consistently better rendering in comparison to Nice-SLAM \cite{niceslam}, Co-SLAM \cite{coslam}, ESLAM \cite{eslam} and BSLAM \cite{birn-slam}. We report the PSNR, SSIM and LPIPS as metrics to reflect the rendering quality. Our model demonstrates advanced results across all metrics. }
\label{tab:per_scene_replica_2d_evaluation}
\end{table*}

\end{document}